\documentclass[journal]{IEEEtran}

\usepackage{xcolor} % for revise

\usepackage{amsmath,amsfonts}
\usepackage{amsthm}
\usepackage{algorithmic}
\usepackage[ruled]{algorithm2e}

\usepackage{mathtools}

\usepackage{xcolor,soul,framed} %,caption
\usepackage{booktabs}
\colorlet{shadecolor}{yellow}
\usepackage{color,soul}
\usepackage[pdftex]{graphicx}

\usepackage{array}
\usepackage[hyphens]{url}
\usepackage{multirow}
\usepackage{amsmath}
\usepackage{threeparttable}

\usepackage{subfigure}
\usepackage{graphicx}
\usepackage{multirow}
\usepackage{hyperref}
\usepackage{bbm}
\usepackage{bm}
\newtheoremstyle{problemstyle}
  {\topsep}   % Space above
  {\topsep}   % Space below
  {}          % Body font (\itshape, \normalfont, etc.)
  {}          % Indent amount
  {\bfseries} % Head font
  {.}         % Punctuation after head
  { }         % Space after head
  {\bfseries\thmname{#1}\thmnumber{ #2}\normalfont\thmnote{ #3}}

\theoremstyle{problemstyle}
\newtheorem{problem}{Problem}

\begin{document}

\title{Orchestrate, Generate, Reflect: A VLM-Based Multi-Agent Collaboration Framework for Automated Driving Policy Learning}

\author{
Zengqi Peng, Yusen Xie, Yubin Wang, Rui Yang, Qifeng Chen,
and Jun Ma, \textit{Senior Member, IEEE}
\thanks{Zengqi Peng, Yusen Xie, Yubin Wang, and Rui Yang are with the Robotics and Autonomous Systems Thrust, The Hong Kong University of Science and Technology (Guangzhou), China (e-mail: zpeng940@connect.hkust-gz.edu.cn; yxie827@connect.hkust-gz.edu.cn; ywang575@connect.hkust-gz.edu.cn; 
ryang253@connect.hkust-gz.edu.cn). }
\thanks{Qifeng Chen is with the Department of Computer Science and Engineering, The Hong Kong University of Science and Technology, Hong Kong SAR, China (e-mail: cqf@ust.hk). }
\thanks{Jun Ma is with the Robotics and Autonomous Systems Thrust, The Hong Kong University of Science and Technology (Guangzhou), Guangzhou 511453, China, and also with the Cheng Kar-Shun Robotics Institute, The Hong Kong University of Science and Technology, Hong Kong SAR, China (e-mail: jun.ma@ust.hk).} 
}

\maketitle

\begin{abstract}

The advancement of foundation models fosters new initiatives for policy learning in achieving safe and efficient autonomous driving. However, a critical bottleneck lies in the manual engineering of reward functions and training curricula for complex and dynamic driving tasks, which is a labor-intensive and time-consuming process. To address this problem, we propose OGR  (\textbf{O}rchestrate, \textbf{G}enerate, \textbf{R}eflect), a novel automated driving policy learning framework that leverages vision-language model (VLM)-based multi-agent collaboration. Our framework capitalizes on advanced reasoning and multimodal understanding capabilities of VLMs to construct a hierarchical agent system. Specifically, a centralized orchestrator plans high-level training objectives, while a generation module employs a two-step analyze-then-generate process for efficient generation of reward-curriculum pairs. A reflection module then facilitates iterative optimization based on the online evaluation. Furthermore, a dedicated memory module endows the VLM agents with the capabilities of long-term memory. To enhance robustness and diversity of the generation process, we introduce a parallel generation scheme and a human-in-the-loop technique for augmentation of the reward observation space. Through efficient multi-agent cooperation and leveraging rich multimodal information, OGR enables the online evolution of reinforcement learning policies to acquire interaction-aware driving skills. Extensive experiments in the CARLA simulator demonstrate the superior performance, robust generalizability across distinct urban scenarios, and strong compatibility with various RL algorithms. Further real-world experiments highlight the practical viability and effectiveness of our framework. 
The source code will be available upon acceptance of the paper.

\end{abstract}


\section{Introduction}

In recent years, the rapid advancement of foundation models and multimodal learning has substantially propelled progress in robotics and autonomous systems across both academia and industry \cite{xu2023multimodal,hanover2024autonomous,han2025multimodal}. These developments provide promising solutions for achieving safe and efficient autonomous driving. Autonomous vehicles (AVs), as a human-centric system with a high safety demand, are required to make appropriate decisions in real time under dynamic conditions \cite{zheng2025safe,huang2025fast}. However, compared to highway scenarios, urban driving environments are characterized by the diversity of driving scenarios and frequent interactions among traffic participants \cite{cai2023closing,li2024interactive}. Various road topologies and distributions of surrounding vehicles (SVs) pose significant challenges to both driving safety and task efficiency. These factors collectively contribute to complicated driving situations. In this sense, enabling AVs to avoid collisions while maintaining high task efficiency in such complex and highly dynamic scenarios has become a critical and pressing research focus.

\begin{figure*}[!htbp]
    \centering
    \includegraphics[trim=0 0cm 0 0.5cm, width=0.8\linewidth]{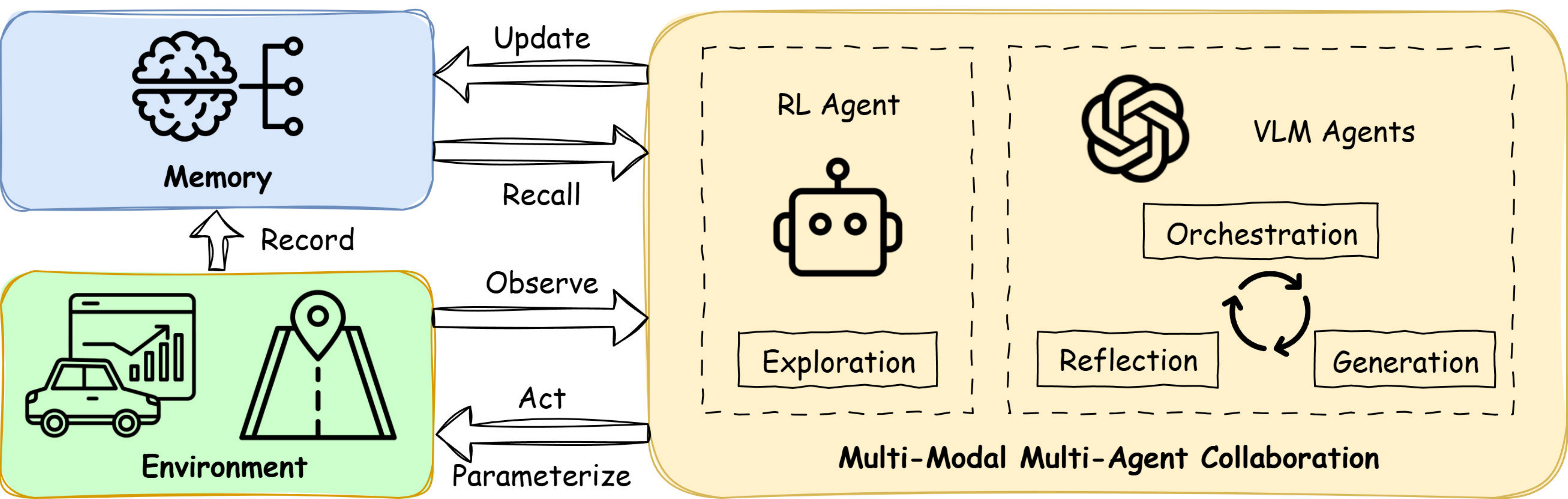}
   \caption{The overview of our VLM-in-the-training-loop paradigm. (1) A VLM-based multi-agent cooperative system interacts with the environment to facilitate the optimization of the RL policy. Specifically, VLM agents collaborate to provide a sequence of reward-curriculum pairs, which are used to parameterize the environment. (2) The RL agent then explores and learns within these configured environments, and the resulting experiences and responses of VLM agents are stored in a memory module. (3) In the next training stage, this memory module provides historical question-answer data and training data to the VLM agents to assist in generating updated reward-curriculum pairs. 
   }
    \label{intro:paradigm}
\end{figure*}
\hspace{0.1cm}

Learning-based approaches for autonomous driving mainly contain imitation learning (IL) and reinforcement learning (RL) \cite{zhu2021survey,chib2023recent,cheng2024pluto}. 
Although IL exhibits high sample efficiency in specific contexts by fitting expert-demonstrated data \cite{liao2025satp}, its performance is fundamentally capped by the quality of the expert data. Furthermore, it often struggles in highly dynamic and interactive urban driving tasks, even with enhancements like DAgger \cite{ross2011reduction} and data augmentation techniques \cite{zare2024survey}. In contrast, RL enables agents to explore and learn through direct trial-and-error interaction with the environment \cite{peng2025bilevel}. It optimizes policies by maximizing cumulative rewards. This interaction-driven process allows RL to adapt to dynamic urban environments and to discover policies that surpass expert performance. However, applying RL to complex urban scenarios introduces two critical challenges \cite{wang2024deep}. The first challenge is to design a reward function that elicits safe and efficient behaviors \cite{biyik2022learning,yu2025reward}. The second one involves ensuring training efficiency, particularly when exploring the vast state-action space \cite{yu2018towards,huang2023efficient}.

Traditionally, researchers have employed techniques like reward shaping \cite{hu2020learning,devidze2022exploration} and curriculum learning (CL) \cite{narvekar2020curriculum,wang2022surveycl} to mitigate these challenges. However, these methods rely heavily on manual engineering, and their effectiveness is ultimately constrained by the prior knowledge and intuition of human experts. The recent rise of large language models (LLMs) and vision-language models (VLMs) presents a transformative opportunity to address this gap \cite{zhang2024vision,yin2024survey}. These foundation models exhibit remarkable emergent abilities that approach the level of comprehensive human expertise, such as powerful multimodal understanding \cite{tiandrivevlm}, common-sense reasoning \cite{sima2024drivelm}, and even code generation \cite{guo2024redcode,wang2024voyager}. Crucially, this capacity for sophisticated reasoning and action allows them to serve as powerful autonomous agents, making collaborative multi-agent systems newly feasible for complex real-world tasks \cite{li2023camel,hu2025owl,hong2024metagpt}. The above advancements position LLMs and VLMs as ideal candidates for automating key components of RL training pipelines, particularly in safety-critical domains like urban autonomous driving.

Based on the above background, our central insight is that the integrated capabilities of advanced VLMs can empower a multi-agent collaborative framework for the automated joint design of reward functions and training curricula. We propose a framework named OGR  (\textbf{O}rchestrate, \textbf{G}enerate, \textbf{R}eflect) for urban driving tasks, while retaining a compatible interface for human expert cooperation. To the best of our knowledge, this work is the first to introduce a collaborative framework leveraging multiple VLM-empowered agents for the automated generation of reward-curriculum pairs in urban autonomous driving scenarios. Fig. \ref{intro:paradigm} depicts the overall architecture of the proposed training paradigm. Our contributions are summarized as follows:

\begin{itemize}
    \item An innovative automated RL policy learning framework is developed for interaction-aware urban autonomous driving. This framework significantly improves the sample efficiency of the RL policy and enhances driving safety in diverse traffic conditions.

    \item A novel VLM-in-the-training-loop paradigm is developed to generate highly consistent and efficient reward-curriculum pairs online. It is composed of multiple VLM agents, including an orchestrator, generators, and a policy reflector. These VLM agents cooperate by engaging in role-playing and coordinated task division, thereby boosting the training efficiency and productivity.

    \item A unique generation module is devised for robust and diverse synthesis of reward-curriculum pairs. It integrates a parallel generation scheme to mitigate VLM hallucinations and a human-in-the-loop mechanism to progressively expand the reward observation space.

    \item Extensive experiments are conducted across both the high-fidelity simulator and hardware platforms. 
    Simulation results demonstrate the superior performance of our framework in terms of task success rate, generalizability, and compatibility across different driving scenarios and RL algorithms. Real-world experiment results further verify the effectiveness of our framework.

\end{itemize}

The rest of the paper is structured as follows. Section II presents the related work. Section III illustrates important preliminaries. Section IV provides a detailed introduction to the proposed framework. Section V includes the experimental validation and corresponding discussion. Finally, Section VI presents the conclusion of this work and outlines potential directions for future work.

\section{Related Work}

\subsection{Deep Reinforcement Learning for Autonomous Driving}

Deep RL has emerged as a mainstream paradigm for learning complex driving policies. However, its successful application in autonomous driving is often hindered by key challenges of reward design and training efficiency \cite{zhao2024surveyad}. Regarding reward design, traditional approaches typically rely on manually engineering the reward function through a process of trial and error \cite{al2023self,wang2023learning}. Although various reward embedding techniques have been proposed to enhance policy performance \cite{chen2022interpretable,chen2024risk}, the multi-objective nature of the driving goals makes this process exceptionally challenging \cite{abouelazm2024review}. For instance, an effective driving policy should simultaneously balance safety, efficiency, and comfort \cite{golchoubian2024uncertainty}. These partly conflicting objectives with varying priorities could make designs based purely on expert intuition laborious and time-consuming, even yielding suboptimal results. As an alternative, Inverse RL is used to extract an intrinsic reward model from expert demonstrations \cite{huang2023conditional,li2024simulation}. But these approaches are constrained by the availability of large-scale, high-quality datasets, and the learned reward models often exhibit poor generalization. In terms of training efficiency, Model-based RL aims to reduce interaction costs by learning an environment model to aid in decision-making or to generate synthetic data \cite{guan2020centralized}. However, the core drawback of this method is its susceptibility to compounding model errors, with performance being limited by the distribution of its training data. On the other hand, CL has proven effective at guiding policy exploration in vast state-action spaces \cite{song2021autonomous,khaitan2022state,peng2023CPPO}. Nevertheless, for complex tasks, predefined curricula that rely on expert knowledge often have limited effectiveness. While adaptive curriculum methods have been proposed \cite{peng2024reward,peng2025bilevel}, they typically rely on assumptions about the structured properties of the tasks and require extensive parameter tuning when applied to new scenarios.

\subsection{Foundation Models for Autonomous Driving}

In recent years, powerful foundation models have been widely applied to numerous frontier domains, including code generation, robotics, and autonomous driving. A direct integration paradigm in autonomous driving is to employ these models as a functional component within the onboard system, such as for perception, prediction, or decision-making \cite{mei2024continuously,xu2024drivegpt4}. LMDrive \cite{shao2024lmdrive} directly generates control signals for AVs by processing multimodal sensor data. Dilu \cite{wen2023dilu} leverages LLMs to enable the system to perform decision-making by customized reasoning and reflection modules. However, the high inference latency inherent to LLMs makes it difficult to satisfy the stringent real-time requirements of driving tasks. To address this problem, DriveVLM-Dual \cite{tiandrivevlm} proposed a fast-slow system that pairs a VLM with a traditional planner in parallel. Although this design mitigates the latency issue, the overall performance bottleneck of the system shifts to the traditional planner acting as the fast component.

On the other hand, foundation models are used to assist the optimization process of RL policies during the training phase. In the area of reward design, existing works like LORD \cite{ye2024lord} and VLM-RL \cite{huang2024vlm} utilize LLMs and VLMs to compute reward signals in real time. However, the dense invocation of LLMs and VLMs can lead to slow training speed and high costs. Recent works, such as Eureka \cite{ma2024eureka}, AutoReward \cite{han2024autoreward}, and SDS \cite{li2024sds}, have explored using LLMs and VLMs to design entire reward functions iteratively, even deriving reward functions from demonstration videos. For training environment and curriculum design, studies such as Eurekaverse \cite{liang2024eurekaverse}, CurricuLLM \cite{ryu2024curricullm}, and CurricuVLM \cite{sheng2025curricuvlm} have demonstrated the ability of these models to generate increasingly challenging environments, decompose complex tasks, and create safety-critical scenarios. However, the majority of these methods focus on automating only a single aspect of the pipeline, either the reward or the curriculum. While the recent LearningFlow framework \cite{peng2025learningflow} explores the generation of both aspects, its design suffers from two key limitations. First, it relies solely on numerical and textual feedback, neglecting rich and intuitive information from multimodal data such as images and videos. Second, its reward and curriculum generation processes are independent. This lack of coordination can potentially lead to a mismatch between the reward function and the curriculum, which could compromise the stability of policy training and degrade the performance of the trained policy.

\section{Preliminaries}
\label{Section2}
\subsection{Task Definition}

This study aims to address the challenge of training an RL policy to achieve safe, efficient, and interaction-aware performance across various complex urban driving tasks, such as multi-lane overtaking, on-ramp merging, and unsignalized intersections. Within our defined task formulation, the ego vehicle (EV) and SVs are initialized at arbitrary, yet traffic-rule-compliant, positions with varying traffic densities. The behavior of EV is determined by an RL policy, while the driving styles of the SVs are randomly initialized to represent a heterogeneous mix of driver behaviors. Under such settings, the RL policy is required to guide the EV to safely and efficiently complete the driving task in environments characterized by diverse configurations and traffic conditions. This task formulation introduces significant stochasticity and high dynamic complexity into the task scenarios, making the task scenarios closer to those in the real world and thus presenting greater challenges.

\subsection{Learning Environment}

We formulate the target tasks as a Markov Decision Process (MDP), which can be represented as a tuple $\langle \mathcal{S}, \mathcal{A}, \mathcal{P}, \mathcal{R}, \gamma \rangle$. The definitions of all components are introduced as follows:

\textbf{State space $\mathcal{S}$}: $\mathcal{S}$ contains kinematic features of vehicles. The state matrix at $k$-th time step is defined as follows:
\begin{equation}
\begin{split}
\mathbf{S}_{k}=\left[\ \mathbf{s}_{k}^0\ \ \mathbf{s}_{k}^1\ ...\ \mathbf{s}_{k}^{N_{\text{sv}}^{\max}}\ \right]^T,
\end{split}
\label{state martix}
\end{equation}
where $N_{\text{sv}}^{\max}$ is the maximum number of observed SVs; $\mathbf{s}_{k}^0$ and $\mathbf{s}_{k}^i\ (i=1,2,...,N_{\text{sv}}^{\max})$ are the states of the EV and the $i$-th SV, respectively. Specifically, $\mathbf{s}_{k}^{i}={\left[\begin{array}{l l l l l l l}{x_{k}^{i}}&{y_{k}^{i}}&{v_{k}^{i}}&{\psi_{k}^{i}}\end{array}\right]}^{T}$, where $x_{k}^{i}$ and $y_{k}^{i}$ represent the X-axis and Y-axis coordinates, respectively; $v_{k}^{i}$ and $\psi_{k}^{i}$ denote the speed and the heading angle, respectively. 

\textbf{Action space $\mathcal{A}$}: $\mathcal{A}$ is a multi-discrete action space consisting of three sub-action spaces:
\begin{equation}
\begin{split}
\mathcal{A}=\left\{ A_{1},A_{2},A_{3} \right\},
\end{split}
\label{Action_space}
\end{equation}
where $A_{1}, A_{2}$, and $A_{3}$ represent three sub-action spaces. The definitions of the sub-action spaces will be detailed in Section \ref{method:RL executor}.

\textbf{State transition dynamics $\mathcal{P}(\mathbf{S}_{k+1}|\mathbf{S}_{k},a_{k})$}: $\mathcal{P}$ determines the probabilities regarding the state transitions which satisfy the Markov property. The environment implicitly defines the probabilities, which are not directly accessible to the RL agent.

\textbf{Reward function $\mathcal{R}$}: $\mathcal{R}$ encourages appropriate and desirable behaviors of the RL agent, while penalizing incorrect or suboptimal ones. It plays a critical role in guiding the exploration and learning of the RL agent. In this work, we harness the multimodal reasoning capabilities of VLMs to automatically generate and recursively optimize reward functions for the RL agent in complex tasks.

\textbf{Discount factor $\gamma$}: $\gamma \in (0,1)$ is utilized to discount future accumulated rewards.

\subsection{Reward Function Generation}
\label{section2_rfgp}

In this work, the goal of reward function design is to construct a shaped reward function for complex autonomous driving tasks in which the ground-truth reward is difficult to obtain directly. Generally, the ground-truth reward is sparse, making it challenging to learn effective RL policies in high-dimensional, continuous state spaces. Furthermore, the ground-truth reward is usually not directly observable and can only be accessed indirectly through queries. The definition of the formal reward design problem is shown as follows.

\begin{problem}[\cite{ma2024eureka,singh2009rewards} (Reward Function Design Problem)] \label{def:rfdp}
    The reward function design problem can be represented as a tuple $P=\langle  E, \mathcal{R}_M, M, F \rangle$, where $E$ denotes an environmental distribution which is a set of MDPs; $\mathcal{R}_M$ represents the reward function space; $M$ is an RL agent which uses function $r_M \in \mathcal{R}_M$ to optimize its policy; $F$ is a fitness function which provides a scalar evaluation for a rollout trajectory. The goal is to design an optimal reward function $r^*_M$ whose optimized policy can achieve the maximum fitness score over the environmental distribution $E$.
\end{problem}

Based on Problem \autoref{def:rfdp}, this study aims to generate the reward terms of a reward function by introducing the code agent \cite{huang2023agentcoder,guo2024redcode}, which is defined as the reward function generation problem. The objective of this problem is to generate a reward function code that can optimize the RL policy to maximize the fitness function based on a given context. Considering the requirements of our target task, the fitness function can be defined as the task success rate. 

\subsection{Online Task-Level Curriculum Generation}
\label{section2_cgp}

Given the complex target task, particularly in the highly dynamic and interactive environments, we model the RL policy training as a curriculum learning problem. In this work, the goal of curriculum design is to develop a sequential multi-task sequence for an RL agent. Compared to naively training the RL policy in the most complex environment, a suitable curriculum can help to improve the sample efficiency \cite{zhang2020automatic,tao2024reverse}, thereby enhancing driving safety and efficiency. The curriculum design problem is defined as follows.

\begin{problem}[\cite{narvekar2020curriculum} (Task-Level Curriculum Design Problem)] \label{def:cdp}
    A curriculum is a directed acyclic graph. It can be represented by a tuple $\mathcal{C} = \langle \mathcal{V}, \mathcal{E}, g, \mathcal{T} \rangle$, where $\mathcal{V}$ denotes the set of vertices, $\mathcal{E}$ represents the directed edges, g is a function that associates vertices to subsets of all transition samples, and $\mathcal{T}$ denotes a set of tasks. The directed acyclic graph terminates on a single sink node.  
    The goal is to design an optimal curriculum $\mathcal{C}^*=[T^*_1, T^*_2,\cdots, T^*_N]$ that can guide the RL policy to achieve the best performance metrics over the target tasks.
\end{problem}

Based on Problem \autoref{def:cdp}, our objective is to leverage a generative AI agent that generates a training curriculum online, conditioned on contextual information, in order to optimize task-specific performance metrics. The above challenge is referred to as the online task-level curriculum generation problem.

Specifically, the challenge of different urban scenarios can typically be attributed to two key dimensions. For instance, the complexity of overtaking and merging tasks primarily arises from traffic density and the velocity patterns of SVs. For intersection scenarios, it stems from traffic density and the task type, such as turning left, going straight, or turning right. Accordingly, we define the task set for CL as a two-level set as follows:
\begin{equation}
\begin{split}
\mathcal{T}=\{t_{i j} \mid i=0,1, \ldots, N_{\mathrm{l1}}^{\max }, j=0,1, \ldots, N_{\mathrm{l2}}^{\max } \} ,
\end{split}
\label{def:2level_curri_set}
\end{equation}
where $N_{\mathrm{l1}}^{\max }$ and $N_{\mathrm{l2}}^{\max }$ denote the number of discrete values associated with two orthogonal dimensions of task difficulty, respectively. The goal of the online task-level curriculum generation problem can be formulated as follows:
\begin{equation}
\mathcal{C}^*=\{T^*_{n}\}=\arg \max \sum\nolimits \mathcal{R}, 
\label{Equ:goal_otcgp}
\end{equation}
where $\mathcal{C}^*$ denotes the optimal curriculum; $T^*_{n}$ is the optimal training task at the $n$-th training interval; $\mathcal{R}$ represents the reward obtained by the RL agent from the environment.

\subsection{Joint Reward-Curriculum Pair Generation Problem}

The target tasks involve a wide range of environmental configurations, making it challenging for the driving policy to learn appropriate behaviors across diverse situations. To address this challenge, we divide the policy training process into multiple stages and generate a tailored reward-curriculum pair for each stage. Based on Section \ref{section2_rfgp} and Section \ref{section2_cgp}, we define the following problem:

\begin{problem}[(Joint Reward-Curriculum Pair Generation Problem)]
    For target tasks and task set $\mathcal{T}=\{t_{i j} \}$, the driving policy is represented as ${\pi}_{\theta}$. 
    The objective is to generate an optimal reward-curriculum sequence $\mathcal{Q}_{RC}^*=[(R^*_{1}, C^*_1), (R^*_{2}, C^*_2), \cdots, (R^*_{N}, C^*_N)]$ to optimize the RL policy, which can maximize the reward over target tasks $\pi_{\theta}^*=\arg \max _{\mathcal{Q}_{RC}^*} \mathcal{R}$. 
\end{problem}

To improve the performance of the trained policy, it is important to enhance the relationship between the generated reward function and the curriculum. To address this challenge, we will introduce the proposed closed-loop OGR in the next section.

\begin{figure*}[htbp]
    \centering
    \includegraphics[trim=0 0cm 0 0cm, width=0.9\linewidth]{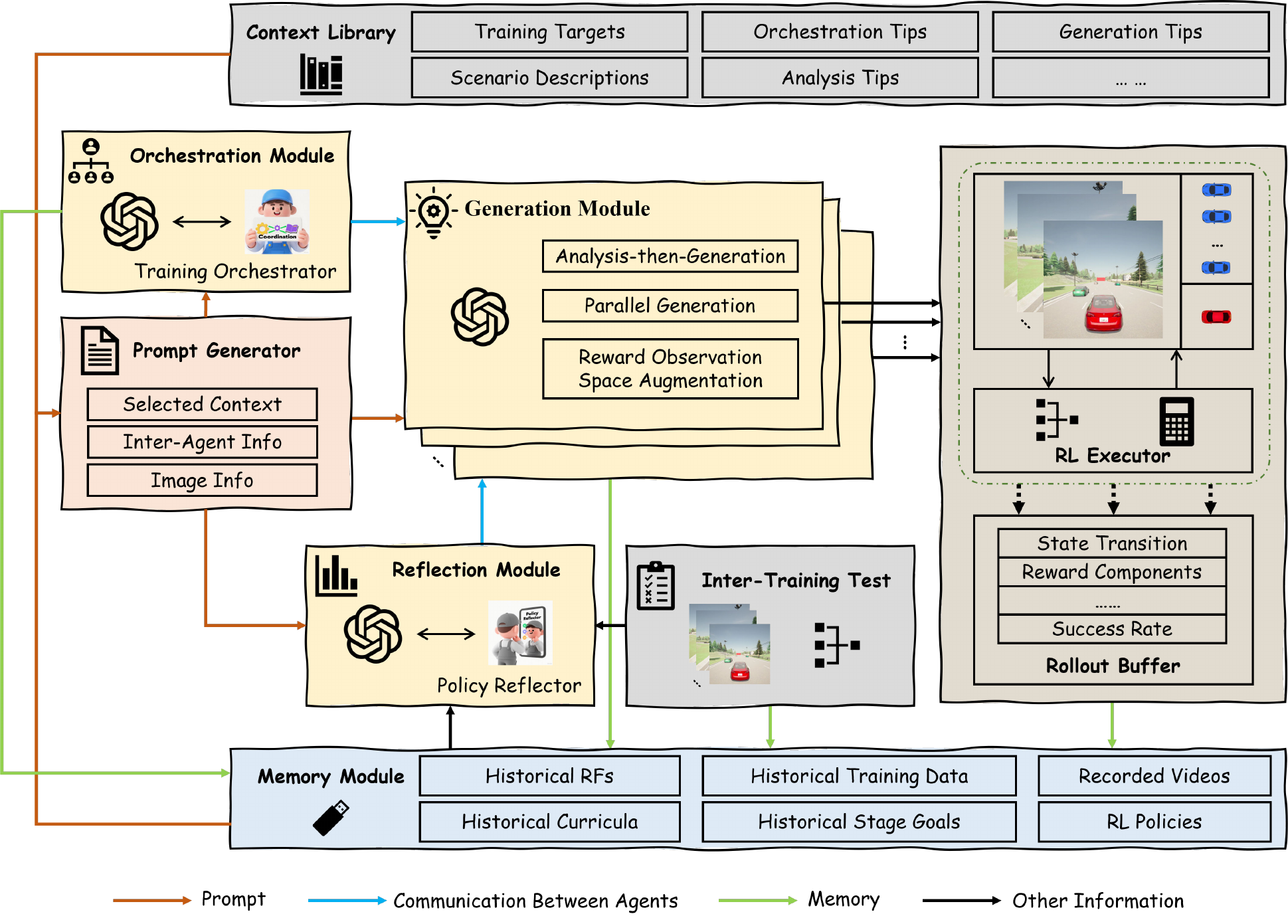}
   \caption{The diagram of the automated RL policy learning framework via VLM-based multi-agent collaboration for autonomous driving tasks. This framework consists of an orchestration module, a generation module, a reflection module, a memory module, a prompt generator, and a context library. 
   }
    \label{frame:proposedframework}
\end{figure*}

\section{Methodology}

\subsection{Overview of Framework}

\begin{figure}[!t]
    \centering
    \includegraphics[trim=0 0cm 0 0cm, width=\linewidth]{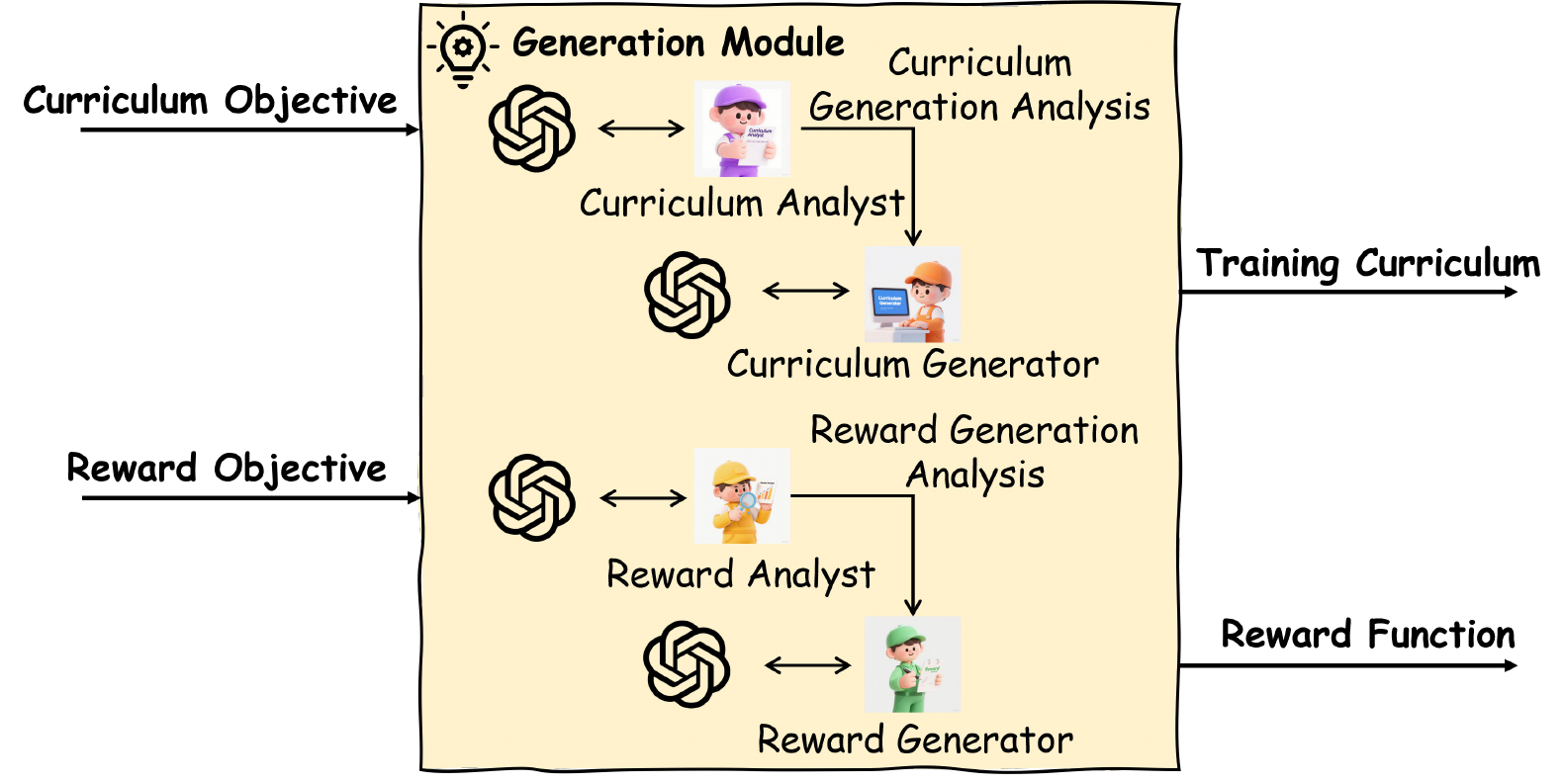}
   \caption{Details of the two-step reward-curriculum generation module.
   }
    \label{frame:2stepgeneration}
\end{figure}

The overall framework is depicted in Fig. \ref{frame:proposedframework}. During the training phase, context is first provided to an orchestrator agent, which formulates a stage-wise training plan. It then issues the training objective of the current stage to the generation module to ensure consistency between the subsequently generated reward function and curriculum. This training objective, combined with the task-specific context, serves as the input for the generation module. Given the complexity of the task scenarios and the generation task, we design this generation process as a two-step procedure, which is shown is Fig. \ref{frame:2stepgeneration}. In the first stage, analysis agents receive this combined information and perform a preliminary analysis of the reward function and curriculum generation. They subsequently output their analytical findings and concrete recommendations. In the second stage, the responses from the analysis agents are delivered to the generation agents along with the generation context. This guides the final generation of specific reward terms and training curricula. Subsequently, the generated reward function and curriculum are passed to the environment module for initialization. They are then used for interaction with the RL agent to collect experience and update its policy. Finally, all responses generated by the VLM agents and the interaction data of the RL agent are recorded in the memory module. 

To enhance training stability, we generate multiple reward-curriculum pairs simultaneously in each training round. After a predefined number of episodes, the current intermediate policy is evaluated in an in-training test. The resulting statistical metrics and video recordings are stored in the memory module and fed to a reflection agent. This agent then selects the optimal reward-curriculum pair to be used in the next training stage. At the end of an orchestration or generation cycle, the results and test data from the previous round are sampled and integrated with the latest contextual information. This consolidated input is then provided back to the corresponding VLM agents for their subsequent reasoning tasks. Notably, during the initial stage of training, only textual context is provided to the VLM agents. Once the training enters the first feedback loop, sampled videos will be integrated as visual input into the contextual information for the VLM agents. The following subsections provide a detailed description.

\subsection{Multimodal Memory Module for Storage and Retrieval}

Mainstream LLMs and VLMs generally lack a built-in long-term memory mechanism. Their operational memory is typically confined to the current context window, preventing them from retaining historical interaction data. However, our proposed framework requires multiple VLM invocations, including orchestration, generation, and reflection, across various stages of a complete RL policy training process. This design necessitates that the reward-curriculum pairs generated by the VLM agents across different training stages maintain a high degree of consistency and temporal coherence.

Therefore, we introduce a memory module specifically for storing multimodal experiences generated throughout the training process. This data primarily includes the interaction history of VLM agents, historical data from interactions between the RL agent and the environment, and video recordings from in-training evaluations. The VLM interaction history further encompasses customized prompts, reasoning process, extracted reward-curriculum pairs, and other valuable information. 
The stored content is selectively retrieved and embedded into the subsequent prompt for the next reasoning round in the form of vectors, text, and image sequences. 
By equipping the framework with this memory module, we endow the VLM agents with long-term memory, thereby enabling the generation of reward-curriculum pairs that exhibit both temporal coherence and logical continuity throughout the training process.

\subsection{High-Level Orchestration for Reward-Curriculum Alignment}

Given the complex target task and the diversity of agent roles, a key challenge in generating reward-curriculum pairs using multiple VLM agents is to achieve effective coordination and orchestration. Prior works usually provide historical query-response data and other context directly and separately to each generation agent, which can lead to a mismatch between the generated reward function and curriculum. This potentially results in inefficient training and suboptimal performance of the final policy. 

Inspired by centralized coordination strategies in multi-agent systems, we introduce a centralized orchestrator agent into our framework. This agent is responsible for the high-level planning of training objectives at different stages of the RL policy training process, dynamically adjusting them based on historical training information. Subsequently, the orchestrator agent provides a clear training objective for the current stage as part of the context for the generation module. 
This centralized design enables the generation module to produce reward-curriculum pairs tailored to the objectives of each specific training stage. This goal-oriented process can result in a significant improvement in training efficiency. Ultimately, this enables the RL policy to acquire driving skills of increasing difficulty more effectively. 

Building on the above description, the orchestration module can be modelled as a two-step process, which can be expressed as follows:
\begin{equation}
G_{n_s^O} = 
\begin{cases} 
\Phi_{VLM}^{O}(l_O,I^{text}) & \text{if } {n_s^O} = 1 \\
\Phi_{VLM}^{O}(l_O,I^{text}, I^{image}, \mathcal{H}_{{n_s^O}-1}) & \text{if } {n_s^O} \ge 2 
\end{cases}
\end{equation}

\begin{equation}
    \left( g_{n_s^O}^{(r)}, g_{n_s^O}^{(c)} \right) = \text{Parse}_O(G_{n_s^O})
\end{equation}
where $n_s^O$ is the index for the training stage; $\Phi_{VLM}^{O}(\cdot)$ and $\text{Parse}_O(\cdot)$ represent the VLM-powered orchestrator agent and parse function, respectively; $l_O$ is the language goal of orchestration; $I^{text}$ and $I^{image}$ denote textual context and visual context, respectively; $G_{n_s^O}$ and $\mathcal{H}_{{n_s^O}-1}$ are the generated comprehensive plan text and multimodal dialogue history, respectively; $g_{n_s^O}^{(r)}$ and $g_{n_s^O}^{(c)}$ represent reward generation objective and curriculum generation objective extracted from $G_{n_s^O}$, respectively.

\subsection{Low-Level Two-Step Reward-Curriculum Generation}

Following the high-level training objective provided by the orchestrator agent, this section details the low-level process of generating specific reward-curriculum pairs. Directly translating a high-level goal into a fine-grained reward function and a complex curriculum sequence is a significant challenge due to the inherent complexity of the target tasks. To address this, we draw inspiration from hierarchical task decomposition in multi-agent systems and reformulate the generation process into two distinct steps: first analysis and then generation. This decoupled approach allows dedicated agents to first interpret and analyze the current training goal and contextual signals before synthesizing the corresponding reward functions and curriculum instances accordingly.

\subsubsection{Pre-Generation Task Analysis for Reward and Curriculum}

Given the heterogeneous nature of the information sources within the multimodal context and the inherent complexity of the generation task, relying on a single agent to perform long-chain reasoning often leads to a significant degradation in performance. Therefore, we introduce two distinct multimodal analysis agents, a reward analysis agent and a curriculum analysis agent. They are responsible for conducting an in-depth analysis of the current RL policy training progress and behavioral patterns based on the multimodal context. Subsequently, they provide separate and specialized guiding blueprints for the reward function and curriculum generation of the next training stage. 
Specifically, these analysis agents leverage the superior capabilities of the underlying VLM for comprehensive multimodal understanding and detailed description generation. 

To facilitate their specialized roles, each analysis agent is provided with a distinct stream of customized information. For instance, the reward analysis agent receives the stage-wise objective for reward generation from the orchestrator agent, relevant high-level guidelines, and historical data detailing the numerical trajectories of the total reward and its individual components. Its primary function is to leverage these inputs to devise a plan for the reward function, outputting its overall design objectives in a structured format, including specific reward terms, meanings, types, and value ranges. In parallel, the curriculum analysis agent processes its own set of inputs, including the objective for curriculum selection, guidelines for curriculum switching, and historical statistics such as the success, collision, and timeout rates of previously executed curricula. Its role is then to evaluate the efficacy of the prior curriculum and analyze its suitability for the upcoming training stage. Therefore, the analysis process of reward and curriculum can be expressed as follows:
\begin{equation}
A_{n_s^R}^{(r)}=  \left\{\begin{array}{l}
\Phi_{VLM}^{(r,A)}\left(l_{(r,A)},g_1^{(r)},I_{(r,A)}^{text}\right), \\ 
\hfill  \text { if } {n_s^R}=1, \\ \Phi_{VLM}^{(r,A)}\left(l_{(r,A)},g_{n_s^R}^{(r)},I_{(r,A)}^{text}, I_{(r,A)}^{image}, \mathcal{H}_{{n_s^R}-1}^{(r,A)}\right), \\ 
\hfill \text { if } {n_s^R} \geq 2,
\end{array} \right.
\end{equation}
\begin{equation}
A_{n_s^C}^{(c)}= \left\{\begin{array}{l}
\Phi_{VLM}^{(c,A)}\left(l_{(c,A)},g_1^{(c)},I_{(c,A)}^{text}\right), \\ 
\hfill \text { if } {n_s^C}=1, \\ \Phi_{VLM}^{(c,A)}\left(l_{(c,A)},g_{n_s^C}^{(c)},I_{(c,A)}^{text}, I_{(c,A)}^{image},\mathcal{H}_{{n_s^C}-1}^{(c,A)}\right), \\ 
\hfill \text { if } {n_s^C} \geq 2,
\end{array} \right.
\end{equation}
where $\Phi_{VLM}^{(r,A)}$ and $\Phi_{VLM}^{(c,A)}$ denote VLM-powered analysis agent of reward and curriculum, respectively; $n_s^R$ and $n_s^C$ are the indices for the generation stage of reward and curriculum, respectively; $l_{(r,A)}$ and $l_{(c,A)}$ represent the language goals of reward and curriculum analysis, respectively; $I_{(r,A)}^{text}, I_{(r,A)}^{image}, I_{(c,A)}^{text}$, and $I_{(c,A)}^{image}$ denote textual context and visual context of reward and curriculum analysis, respectively; $\mathcal{H}_{{n_s^R}-1}^{(r,A)}$ and $\mathcal{H}_{{n_s^C}-1}^{(c,A)}$ are multimodal dialogue history of reward and curriculum analysis, respectively; $A_{n_s^R}^{(r)}$ and $A_{n_s^C}^{(c)}$ denote in-depth structured analysis of reward and curriculum.

\subsubsection{Parallel Reward-Curriculum Generation and Human-in-the-Loop Reward Observation Space Augmentation}

Building upon the structured analysis provided by the analysis agents, the generation agents proceed to generate the reward functions and training curricula. To ensure this process is systematic, consistent, and produces high-quality outputs, we employ a comprehensive prompt template. The construction of this template begins with role assignment, which establishes the identity of agents by positioning them as a specialized generator for a given task. Next, during context injection, we provide key context information, including the available environment observation variable code, the definitions and contents of the predefined curriculum set, and the fine-grained guidelines produced by the analysis agents. Finally, in the memory integration step starting from the second round onward, the complete dialogue history from the previous round is supplied as a long-term memory context. The VLM-based agents then reference this context to inform their subsequent inference.

Within this prompting framework, two generation agents execute their respective tasks. 
The core responsibility of the reward generation agent is to translate the analytical blueprint from the preceding step into a concrete code implementation. Its workflow begins with a deliberate evaluation of each reward term proposed in the analysis. During this evaluation, the agent comprehensively reviews relevant environment observation variables and historical reward functions. It then decides on a component-by-component basis whether to include each term and designs its precise mathematical expression. Finally, all selected components are aggregated into an executable reward function code. Meanwhile, the curriculum generation agent focuses on the dynamic adjustment of the curriculum sequence. It leverages the analysis conclusion and training progress context to develop a comprehensive understanding of the current learning state of the policy. Subsequently, it initiates a situational assessment procedure, comparing historical training data against the task settings of the previous stage to judge if the curriculum difficulty is well-matched to the capabilities of the current policy. Based on this assessment, the agent then selects the most suitable set of tasks from the predefined curriculum set for the upcoming training stage. Therefore, the generation process of reward and curriculum can be expressed as follows:
\begin{equation}
R_{n_s^R}^{(G)} = \left\{\begin{array}{l}
\Phi_{VLM}^{(r,G)}\left(l_{(r,G)}, A_1^{(r)}, I^{text}_{(r,G)}\right), \\ 
\hfill  \text{if } {n_s^R} = 1, \\
\Phi_{VLM}^{(r,G)}\left(l_{(r,G)}, A_{n_s^R}^{(r)}, I^{text}_{(r,G)}, I^{image}_{(r,G)}, \mathcal{H}_{{n_s^R}-1}^{(r,G)}\right), \\ 
\hfill  \text{if } {n_s^R} \ge 2 ,
\end{array} \right.
\end{equation}
\begin{equation}
    R_{n_s^R} = \text{Parse}_R\left(R_{n_s^R}^{(G)}\right),
\end{equation}
\begin{equation}
C_{n_s^C}^{(G)} = \left\{\begin{array}{l}
\Phi_{VLM}^{(c,G)}\left(l_{(c,G)}, A_1^{(c)}, I^{text}_{(c,G)}\right), \\ 
\hfill  \text{if } {n_s^C} = 1, \\
\Phi_{VLM}^{(c,G)}\left(l_{(c,G)}, A_{n_s^C}^{(c)}, I^{text}_{(c,G)}, I^{image}_{(c,G)}, \mathcal{H}_{{n_s^C}-1}^{(c,G)}\right), \\ 
\hfill  \text{if } {n_s^C} \ge 2, 
\end{array} \right.
\end{equation}
\begin{equation}
    C_{n_s^C} = \text{Parse}_C\left(C_{n_s^C}^{(G)}\right),
\end{equation}
where $\Phi_{VLM}^{(r,G)}$ and $\Phi_{VLM}^{(c,G)}$ denote VLM-powered generation agent of reward and curriculum, respectively; $l_{(r,G)}$ and $l_{(c,G)}$ represent the language goals of reward and curriculum generation, respectively; $I_{(r,G)}^{text}, I_{(r,G)}^{image}, I_{(c,G)}^{text}$, and $I_{(c,G)}^{image}$ denote textual context and visual context of reward and curriculum generation, respectively; $\mathcal{H}_{{n_s^R}-1}^{(r,G)}$ and $\mathcal{H}_{{n_s^C}-1}^{(c,G)}$ are multimodal dialogue history of reward and curriculum generation, respectively; $R_{n_s^R}^{(G)}$ and $C_{n_s^C}^{(G)}$ denote generated content of reward and curriculum; $\text{Parse}_R(\cdot)$ and $\text{Parse}_R(\cdot)$ represent the parse function of reward function and curriculum, respectively; $R_{n_s^R}$ and $ C_{n_s^C}$ are extracted reward function code and curriculum index, respectively.

Once the executable reward function code and selected curriculum identifiers are parsed from the raw output of the VLM, these components are passed to the RL training environment via a standardized interface. Their primary role is to configure and parameterize the environment for the interaction of the RL agent and experience collection in the upcoming training stage.

To mitigate the potential impact of VLM hallucinations on inferred content quality, our framework generates a set of reward-curriculum pairs in parallel within each training stage. These pairs are subsequently used to parameterize multiple environment instances for the exploration of the RL agent, and the interaction history from each instance is independently logged to the memory module along with its corresponding generated pair. This parallel generation can improve the safety and reliability of the generations through diversity, and ultimately enhance the performance of the final trained policy.

\subsubsection{Human-in-the-Loop Reward Observation Space Augmentation}
Notably, the environment observation code within our framework is designed to be extensible as training progresses. Initially, it is constructed based on the prior knowledge of human experts in reward design. However, this initial observation library is only a subset of the full range of potential observations. To dynamically expand this library, we introduce a dedicated human-in-the-loop mechanism for reward observation space augmentation. Under this mechanism, if the expression of a reward component needs an observation variable absent from the current observation library, the proposal will not be dismissed or result in a programming error. Instead, it will be output in a structured format and flagged for human expert review. Before the next training round, the human expert assesses the feasibility and potential utility of the proposed variable. If it is accessible and beneficial, the expert will expose this new variable within the environment observation code. This augmentation process allows the observation code to co-evolve with the reward function and RL policy, thereby potentially enhancing the quality of the generated reward functions.

\subsection{Automated Policy Evolution via In-Training Reflection}

Upon completion of a training stage, where the RL agent has finished a set of episodes across multiple environments parameterized by $N_G$ parallel-generated reward-curriculum pairs, the resultant intermediate policy $\left\{\pi_{n_s^G}^{(i)} \mid i=1,2,\cdots,N_G \right\}$ from each branch is extracted for an in-training test. Specifically, the test requires each intermediate policy to perform repeatedly in the most challenging task setting from the curriculum it has just experienced. Key performance metrics and the full video recordings from this testing process are archived in the memory module.

To conduct a comprehensive evaluation of the actual performance of the policy that goes beyond traditional metrics, we additionally introduce a scoring agent. This agent receives $N_S$ key clips $\left\{ I_j^{(i)} \mid j=1,2, \cdots,N_S \right\}$ sampled from the test videos and, based on a meticulously designed and structured scoring prompt, outputs a quantitative behavioral score. Subsequently, the average behavioral score, along with the aforementioned statistical metrics, is delivered to a reflection agent. The core task of the Reflection Agent is to assess the training effectiveness of each reward-curriculum pair within the past stage. Ultimately, it selects the best-performing reward-curriculum pair and its corresponding intermediate policy from all parallel branches. This selected combination then serves as the foundation for the next training stage, informing the new round of reward-curriculum generation and the subsequent update of the RL policy. Therefore, the complete reflection process can be expressed as follows:
\begin{equation}
    b_j^{(i)} = \text{Parse}_S\left( \Phi_{VLM}^{S}\left(l_{S},I_j^{(i)}\right) \right),
\end{equation}
\begin{equation}
    \bar{B}_{n_s^O}^{(i)} = \frac{1}{N_S} \sum_{j=1}^{N_S} b_j^{(i)},
\end{equation}
\begin{equation}
    E_{n_s^O} = \Phi_{VLM}^{R}\left(l_{R},\left\{M_{n_s^O}^{(i)}, \bar{B}_{n_s^O}^{(i)}\right\}_{i=1}^{N_G}\right),
\end{equation}
\begin{equation}
    i^*_{n_s^O} = \text{Parse}_R\left( E_{n_s^O}\right),
\end{equation}
where $\Phi_{VLM}^{S}$ and $\Phi_{VLM}^{R}$ are VLM-powered scoring agent and reflection agent, respectively; $\text{Parse}_S\left( \cdot \right)$ and $\text{Parse}_R\left( \cdot \right)$ represent the parse function of scoring and reflection, respectively; $l_{S}$ and $l_{R}$ denote the prompt instruction of scoring and reflection, respectively; $ b_j^{(i)}$ represent the visual behavioral score for $j$-th clip of RL policy $\pi_{N_s^G}^{(i)}$; $\bar{B}_{n_s^O}^{(i)}$ is the average behavioral score; $M_{n_s^O}^{(i)}$ represents statistical metrics; $E_{n_s^O}$ and $i^*_{n_s^O}$ denote the response of the reflection agent and parsed index of the optimal policy in training stage $n_s^O$, respectively.

\subsection{Downstream RL Executor}
\label{method:RL executor}

After the training environments are configured by the multiple reward-curriculum pairs, a hierarchical RL-model predictive control (RL-MPC) architecture \cite{peng2025bilevel} is used as the downstream executor for interaction. Specifically, the RL policy is implemented as a neural network $\pi$ with parameters $\bm{\theta}$. This policy processes observations to determine a set of decision variables, which are then used to direct the model predictive controller (MPC) to generate low-level control signals. Therefore, given the time step $k$, the action of the RL agent is derived from the current observation $\mathbf{O}_k$ as follows:
\begin{equation}
    a_k^{RL} = \pi_{\bm{\theta}}(\mathbf{O}_k)
    \label{eq:get_action}
\end{equation}
where the observations are defined as follows:
\begin{equation}
\begin{split}
\mathbf{O}_{k} &= \left[\ \mathbf{o}_{k}^0\ \ \mathbf{o}_{k}^1\ ...\ \mathbf{o}_{k}^{N_{\text{obs}}^{\max}}\ \right]^T \\
&= \begin{bmatrix}
  {\
\Delta x_{k}^{0}}&{\Delta y_{k}^{0}}&{v_{k}^{0}}&{\Delta \psi_{k}^{0}} \\
  {\
\Delta x_{k}^{1}}&{\Delta y_{k}^{1}}&{\Delta v_{k}^{1}}&{\Delta \psi_{k}^{1}} \\
\vdots & \vdots & \vdots & \vdots \\
  {\
\Delta x_{k}^{N_{\text{obs}}^{\max}}}&{\Delta y_{k}^{N_{\text{obs}}^{\max}}}&{\Delta v_{k}^{N_{\text{obs}}^{\max}}}&{\Delta \psi_{k}^{N_{\text{obs}}^{\max}}} \\
\end{bmatrix},
\end{split}
\label{RL_obs_matrix}
\end{equation}
where $N_{\text{obs}}^{\max}$ is the maximum number of the observable SVs; $\Delta x_{k}^{i}, \Delta y_{k}^{i},\Delta v_{k}^{i},$ and $\Delta \psi_{k}^{i}$ represent the respective deviations of the EV from the target pose ($i=0$) and from the $i$-th SV ($i=1,2,\dots,N_{\text{obs}}^{\max}$).

The multi-discrete action space (\ref{Action_space}) consists of three sub-action spaces, including waypoint, reference velocity, and lane change sub-action spaces. The mathematical definition is expressed as follows:
\begin{equation}
    A_1 = \left\{ \textup{WP}_0, \textup{WP}_1, ..., \textup{WP}_4\right\}, 
    \label{AS1}
\end{equation}
\begin{equation}
    A_2 = \left\{ 0,\frac{v_{limit}}{4},\frac{v_{limit}}{2},\frac{3v_{limit}}{4},v_{limit}\right\}, 
    \label{AS2}
\end{equation}
\begin{equation}
    A_3 = \left\{ -1,0,1 \right\},
    \label{AS3}
\end{equation}
where $\textup{WP}_i= [x^{\text{WP}}_i\ y^{\text{WP}}_i\ \psi^{\text{WP}}_i]^T$ includes the related reference information of the $i$-th waypoint; $v_{limit}$ is speed limitation of the current road; $-1,0$, and $1$ denote the left lane-change decision, lane-keeping decision, and right lane-change decision, respectively. Waypoints are predefined by the road map. Before the task commences, a reference waypoint set is generated by the $A^*$ search algorithm. At each time step, the five waypoints nearest to the EV are selected to construct the $A_1$. The synergy of these three sub-action spaces endows the EV with a rich repertoire of motion patterns, which is critical for navigating complex interactions with SVs exhibiting heterogeneous behaviors. After the RL policy outputs decision variables, they are first decoded into a set of concrete parameters and then passed to the MPC. Then the MPC performs real-time trajectory planning and computes the control commands applied to the EV.

Relevant training data during the interaction process of the RL agent is stored in the replay buffer. After completing a predetermined count of episodes, a gradient update is applied to the RL policy. This update aims to optimize  the objective function $J(\bm{\theta})$ that corresponds to the reward-curriculum pairs, which can be expressed as follows:
\begin{equation}
\bm{\theta}^*=\arg \max _{\bm{\theta},\left\{(R_{1}, C_1), (R_{2}, C_2), \cdots, (R_{N}, C_N)\right\}} J(\bm{\theta}).
\label{equ:ACRL}
\end{equation}

\section{Experiments}

\subsection{Experimental Setup}

To verify the effectiveness of the proposed method, we conduct comprehensive experiments and summarize the results. We first compare the proposed method against both classical baselines and recent state-of-the-art approaches on a multi-lane overtaking task. This scenario also serves as a qualitative demonstration of the effective collaboration among the VLM agents within our framework. We then further validate its effectiveness in more complex autonomous driving scenarios, including on-ramp merging and unsignalized intersections. We also assess the compatibility of our framework with different classic RL algorithms. We conduct all simulations on a computer system equipped with an Intel(R) Core(TM) i9-14900K CPU and an NVIDIA GeForce RTX 3090 GPU, operating on the Ubuntu 18.04 system. All driving tasks are constructed on the CARLA simulator \cite{dosovitskiy2017carla}, which is an open-source high-fidelity simulator. It features a suite of maps with diverse driving scenarios and configurable driving styles for SVs, which are capable of simulating a wide spectrum of driving conditions. All SVs are controlled by the built-in autopilot mode of the CARLA simulator, and each is assigned a random driving style.

The proposed framework involves cooperation among multiple VLM agents and prompts with multimodal context. This requires a VLM with exceptional capabilities for multimodal information understanding and reasoning to build our multi-agent system. Furthermore, our framework invokes the VLM for inference only during the discrete stages of reward-curriculum pair generation. Consequently, the invocation frequency is significantly lower than that of VLM-in-the-task-loop frameworks. Based on these considerations, we select the GPT-4o model as the foundational VLM for constructing the agents within our system. The number of parallel generations is empirically set to 5. Proximal policy optimization (PPO) \cite{schulman2017proximal} is adopted as the default algorithm to update the policy. This choice is primarily motivated by its effective balance between training stability and sample efficiency. Furthermore, its inherent support for the multi-dimensional discrete action space is critical for our task. The policy update is governed by a clipped objective function as follows:
\begin{equation}
J(\bm{\theta})=\mathbb{E}\left[\min \left(\rho(\bm{\theta}) \hat{A}, \operatorname{clip}\left(\rho(\bm{\theta}), 1-\epsilon, 1+\epsilon\right) \hat{A}\right)\right],
\end{equation}
where $\rho(\bm{\theta})$ denotes the probability ratio between the updated policy and the previous policy; $\hat{A}$ represents the estimated advantage; $\epsilon$ is the clipping threshold.

For the following experiments, we implement the PPO policy for the RL executor using an actor-critic architecture. Both the actor and critic networks are constructed as multi-layer perceptrons in PyTorch. The architecture of each network comprises two hidden layers, the first with 256 neurons and the second with 128. We utilize the Adam optimizer \cite{kingma2014adam} for training these networks. For the MPC module, the parameterized optimization problem is solved by employing the IPOPT solver within the CasADi toolbox \cite{andersson2019casadi} and adopting a single-shooting approach. We train all policies for a maximum of 5000 episodes using the proposed OGR framework. Key hyperparameters include a discount factor $\gamma$ of 0.99, a PPO clip parameter $\epsilon$ of 0.2. The learning rates for the actor and critic networks are set to $5 \times 10^{-4}$ and $1 \times 10^{-3}$, respectively, with 50 epochs per update. The update frequency parameters for orchestration, reward generation, and curriculum generation are set to 1000, 1000, and 100, respectively. Therefore, there are a total of 5 training stages. When applying our method to different experimental scenarios, the only modifications made are to the textual descriptions of the driving task characteristics.

Here, we compare the proposed framework against the following baseline and state-of-the-art (SOTA) methods:

\begin{itemize}
    \item Vanilla PPO \cite{schulman2017proximal}: the vanilla PPO baseline is trained directly in the task scenario using a reward function designed by a domain expert.
    \item Curriculum PPO \cite{peng2023CPPO}: we train the PPO policy with an expert-designed curriculum that distributes episodes over a two-level curriculum set, which includes four traffic densities (1:2:2:5 ratio) and three SV speed modes (1:1:3 ratio).
    \item AutoReward \cite{han2024autoreward}: a state-of-the-art method where the policy is trained directly in the task scenario. In this approach, an LLM-generated reward function is iteratively refined after the whole training process. 
    \item LearningFlow \cite{peng2025learningflow}: a SOTA approach that iteratively generates the reward function and curriculum by invoking LLMs during the training process. 
    \item OGR Without Orchestrator Agent (w/o OA): the proposed method without the orchestrator agent, which serves an ablation study for the orchestration module.
\end{itemize}

To ensure a fair comparison, the parameters for the downstream RL executors are held constant across all evaluated methods. 
We use success rate (SR), collision rate (CR), and time-out rate (TOR) as the main evaluation metrics, where a successful episode is defined as the EV reaching the destination without collision and within the predefined maximum time steps.

\begin{table*}[]
\centering
\caption{Performance comparison among different approaches in the multi-lane overtaking task.
}
\label{table:test_res}
\resizebox{\linewidth}{!}{
\begin{tabular}{@{}ccccccccccccc@{}}
\toprule
\multirow{2}{*}{Methods} & \multicolumn{3}{c}{Empty}              & \multicolumn{3}{c}{Low Density}       & \multicolumn{3}{c}{Medium Density}    & \multicolumn{3}{c}{High Density}      \\ \cmidrule(l){2-13} 
                         & SR (\%)       & CR (\%)     & TOR (\%)    & SR (\%)      & CR (\%)     & TOR (\%)    & SR (\%)      & CR (\%)     & TOR (\%)    & SR (\%)      & CR (\%)     & TOR (\%)    \\ \midrule
Vanilla PPO \cite{schulman2017proximal}              & 99           & 1          & 0          & 80          & 20         & 0          & 69          & 31         & 0          & 62          & 38         & 0          \\
Curriculum PPO \cite{peng2023CPPO}           & 100          & 0          & 0          & 92          & 8          & 0          & 83          & 17         & 0          & 77          & 23         & 0          \\
AutoReward (iter=0) \cite{han2024autoreward}      & 99           & 0          & 1          & 73          & 27         & 0          & 60          & 40         & 0          & 48          & 52         & 0          \\
AutoReward (iter=5) \cite{han2024autoreward}      & 100          & 0          & 0          & 85          & 15         & 0          & 76          & 24         & 0          & 70          & 30         & 0          \\
LearningFlow \cite{peng2025learningflow}            & 100          & 0          & 0          & 96          & 4          & 0          & 90          & 10         & 0          & 85          & 15         & 0          \\
OGR (w/o OA)             & 100          & 0          & 0          & 97          & 3          & 0          & 92          & 8          & 0          & 89          & 11         & 0          \\
\textbf{OGR}             & \textbf{100} & \textbf{0} & \textbf{0} & \textbf{99} & \textbf{1} & \textbf{0} & \textbf{95} & \textbf{5} & \textbf{0} & \textbf{93} & \textbf{7} & \textbf{0} \\ \bottomrule
\end{tabular}}
\end{table*}

\begin{figure}[htbp]
\centering
\includegraphics[width=.9\linewidth]{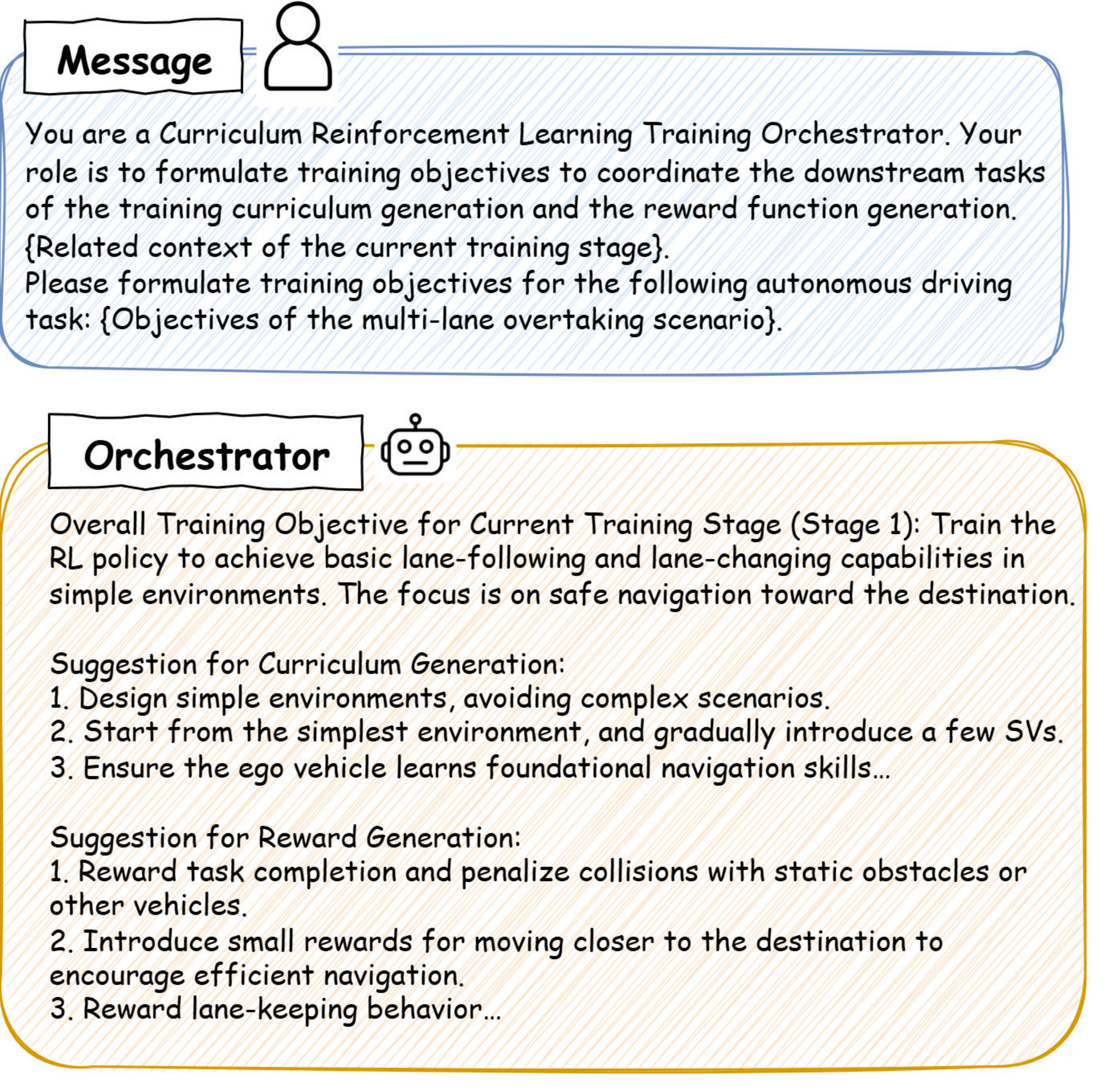}
\caption{A representative sample of the orchestrator.
}
\label{exp:orches}
\end{figure}

\begin{figure}[htbp]
\centering
\includegraphics[width=.9\linewidth]{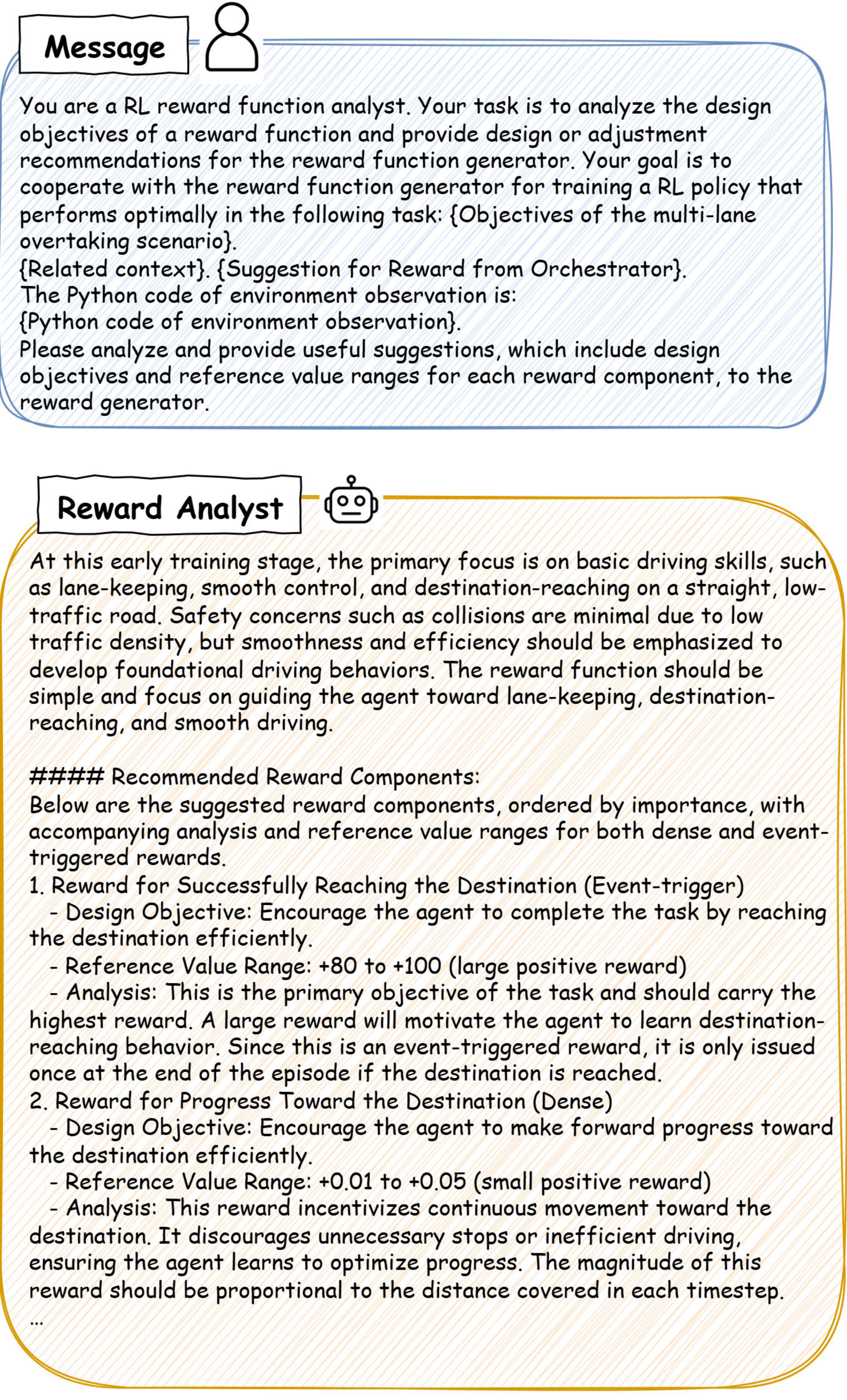}
\caption{A representative sample of the reward analyst.
}
\label{exp:ra}
\end{figure}

\begin{figure}[htbp]
\centering
\includegraphics[width=.9\linewidth]{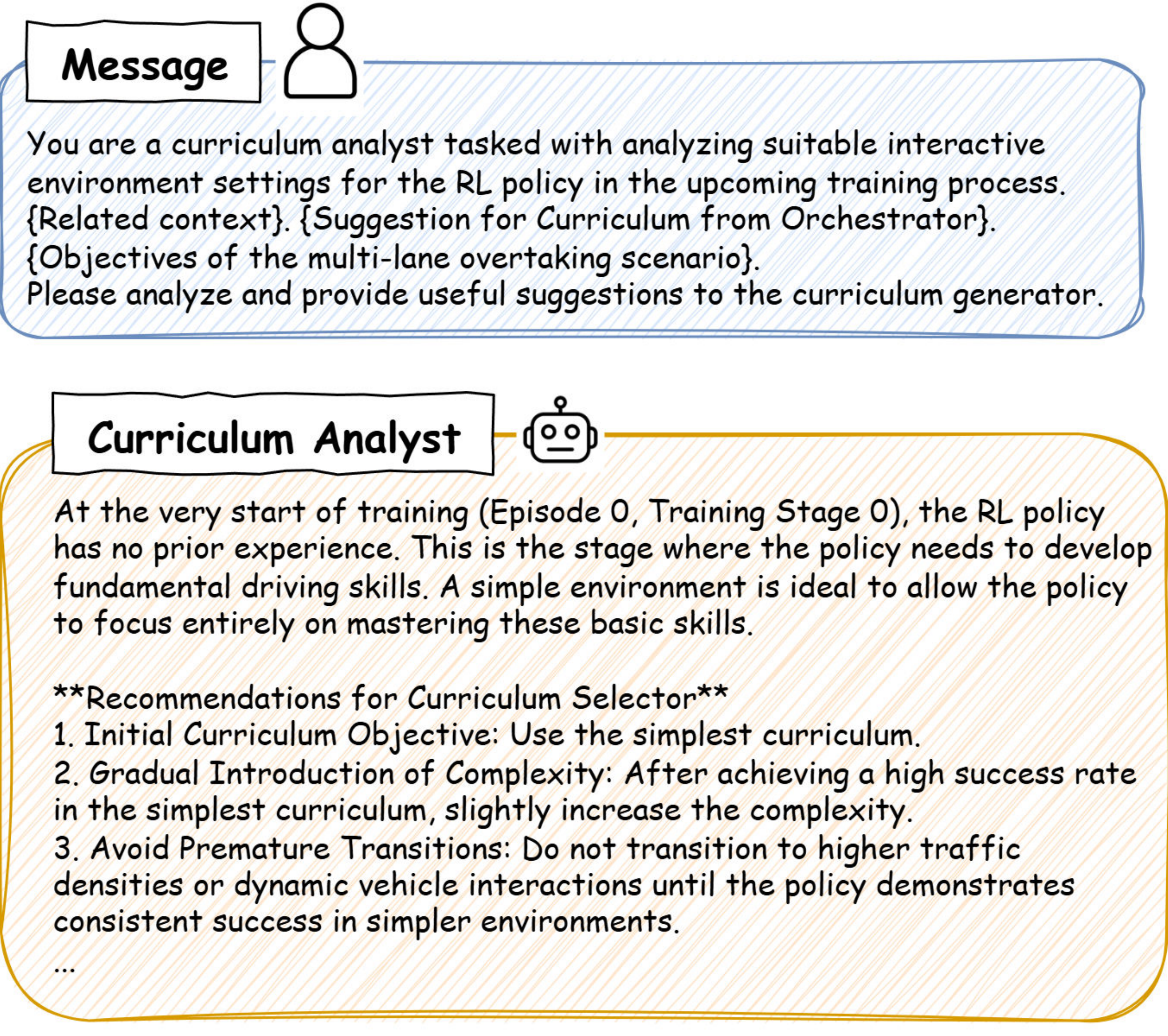}
\caption{A representative sample of the curriculum analyst.
}
\label{exp:ca}
\end{figure}

\begin{figure}[htbp]
\centering
\includegraphics[width=.9\linewidth]{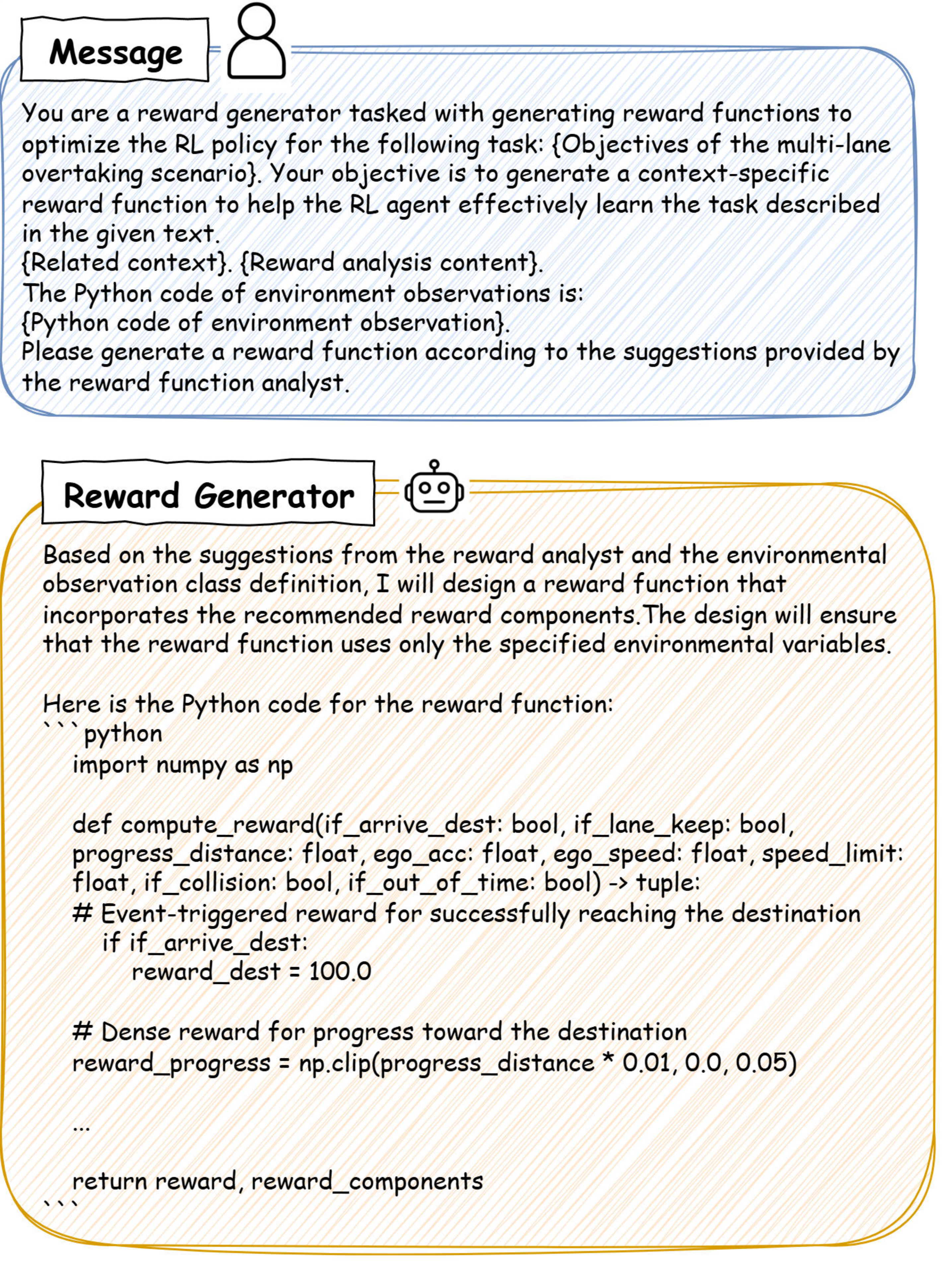}
\caption{A representative sample of the reward generator.
}
\label{exp:rg}
\end{figure}

\begin{figure}[htbp]
\centering
\includegraphics[width=.9\linewidth]{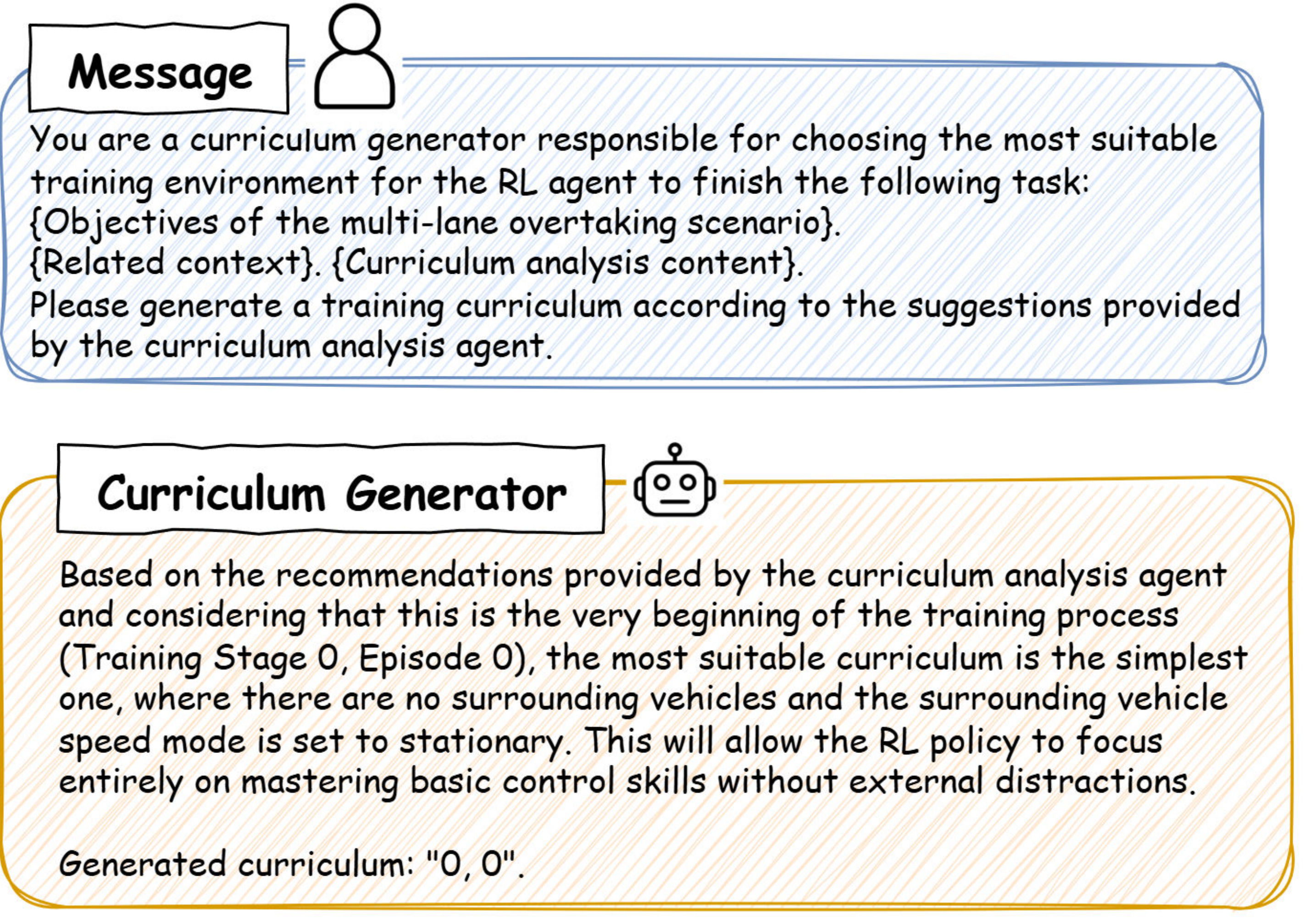}
\caption{A representative sample of the curriculum generator.
}
\label{exp:cg}
\end{figure}

\begin{figure}[htbp]
\centering
\includegraphics[width=.9\linewidth]{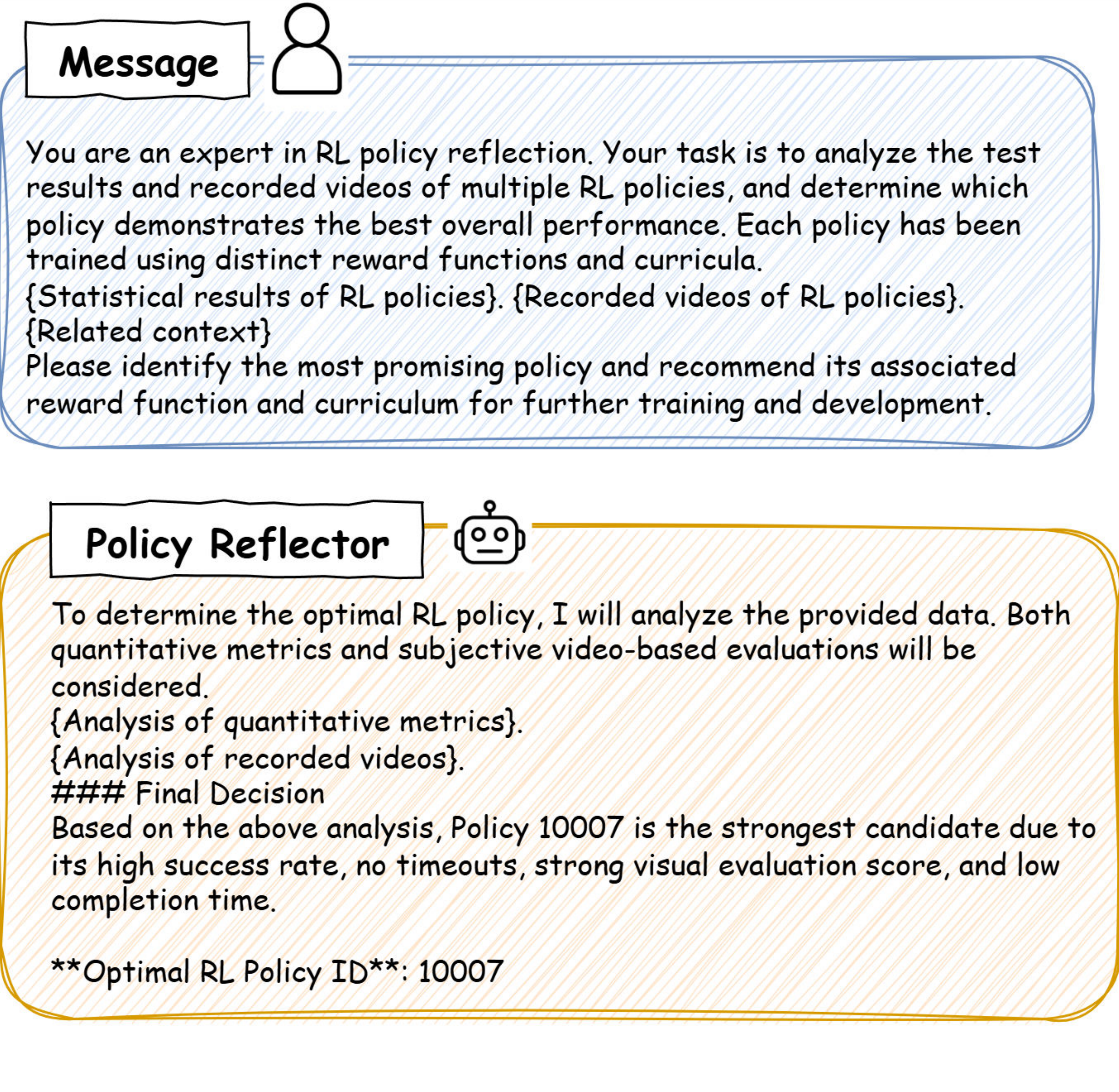}
\caption{A representative sample of the policy reflector.
}
\label{exp:reflect}
\end{figure}

\subsection{Demonstration of Multi-VLM-Agent Collaboration}

To qualitatively demonstrate the effectiveness of our framework, we provide a complete visual example that deconstructs the collaborative workflow among the multiple agents within our framework during a typical training stage. The process begins as shown in Fig. \ref{exp:orches}, which illustrates how the orchestrator establishes a high-level training objective for the current stage and generates initial suggestions for reward and curriculum generation. Subsequently, the analysis phase is depicted in Fig. \ref{exp:ra} and Fig. \ref{exp:ca}, where the reward analyst and curriculum analyst receive corresponding guidance from the orchestrator and perform in-depth analyses by combining it with their unique contextual information. The outputs of this analysis phase are passed to the reward generator and curriculum generator, respectively, to guide the final generation of the reward-curriculum pairs, which are shown in Fig. \ref{exp:rg} and Fig. \ref{exp:cg}.

Within the environments parameterized by these pairs, the RL agent explores, collects experience, and updates the RL policy. After a predetermined number of updates, the multiple intermediate policies from all parallel branches undergo an in-training test. Finally, as shown in Fig. \ref{exp:reflect}, the multimodal results of these tests and corresponding contexts are provided to the policy reflector. Through a comprehensive evaluation, this agent selects the optimal intermediate RL policy, which then serves as the initial policy for the subsequent training stage, thus completing the entire closed loop.

\begin{figure*}[!htbp]
\centering
    \includegraphics[width=0.95\textwidth]{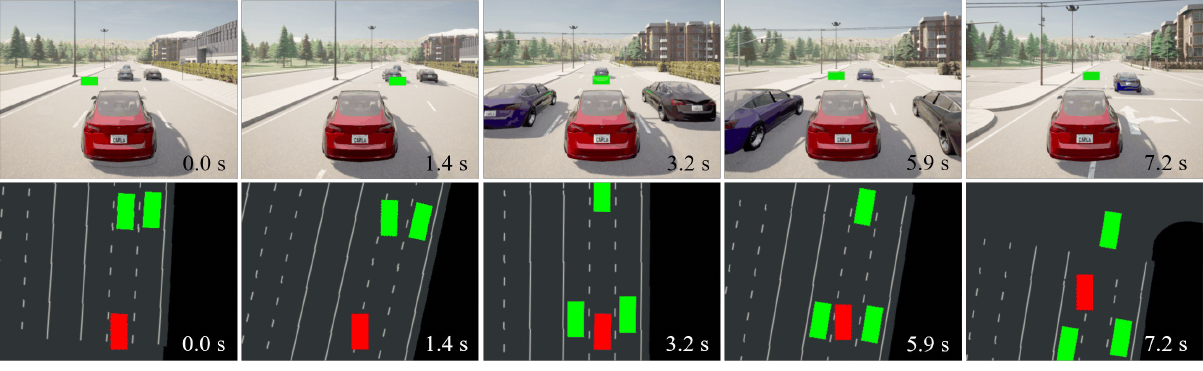}
\caption{Key snapshots of our method in the multi-lane overtaking scenario in the CARLA simulator. The top and bottom of each sub-figure provide third-person views and bird-eye views, respectively. In the third-person view, the green rectangles denote the selected waypoints of the RL policy. In the bird-eye view, the red rectangle represents the EV, while the green rectangles indicate the SVs. 
}
\label{fig:sim_demo_overtake}
\end{figure*}

\begin{figure*}[!htbp]
\centering
    \includegraphics[width=0.95\textwidth]{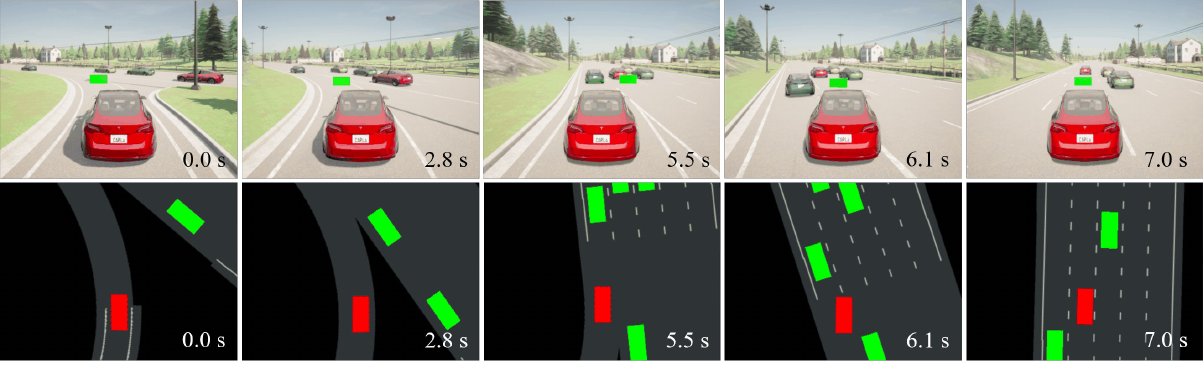}
\caption{Key snapshots of our method in the on-ramp merging scenario in the CARLA simulator. The EV is shown as a red rectangle and the SVs as green rectangles.
}
\label{fig:sim_demo_merging}
\end{figure*}

\begin{figure*}[!htbp]
\centering
    \subfigure[Left-turn task.]{\includegraphics[width=0.95\textwidth]{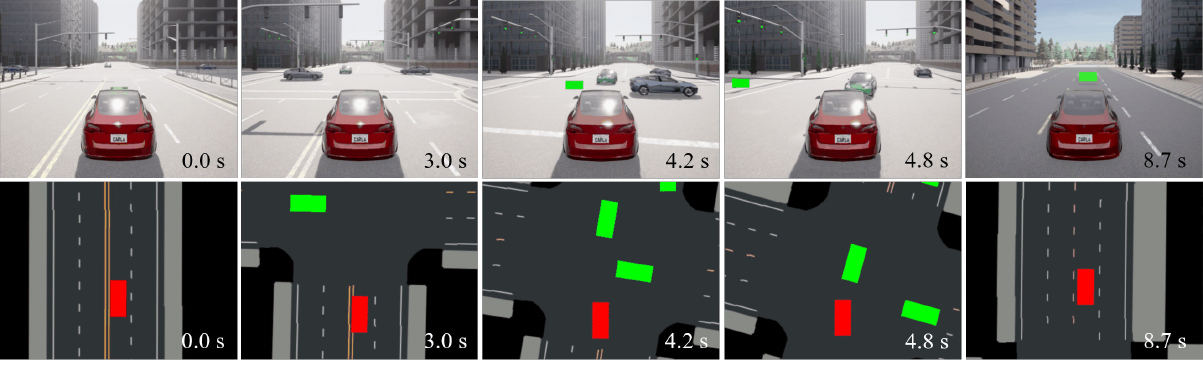}\label{exp:demo_left}}
    \subfigure[Go-straight task.]{\includegraphics[width=0.95\textwidth]{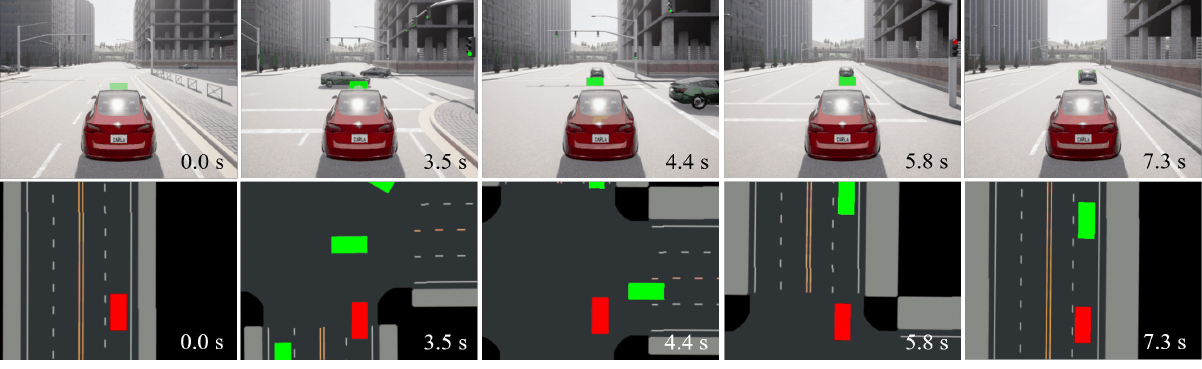}\label{exp:demo_straight}}
    \subfigure[Right-turn task.]{\includegraphics[width=0.95\textwidth]{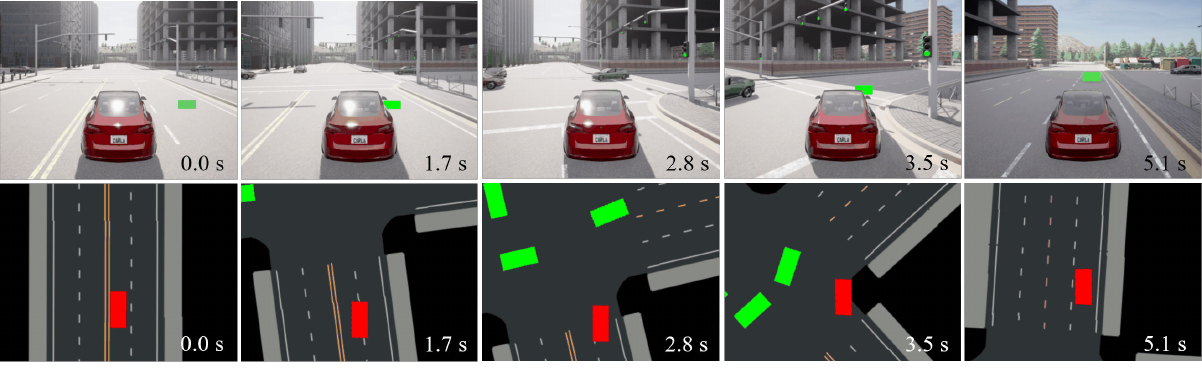}\label{exp:demo_right}}
\caption{Key snapshots of our method in unsignalized intersection scenarios in the CARLA simulator. The EV is shown as a red rectangle and the SVs as green rectangles.
}
\label{fig:sim_demo_intersection}
\end{figure*}

\subsection{Comparative Results and Analysis}

For a quantitative comparison among different methods, we tested the policies trained by our proposed method against several baselines. Specifically, each policy trained on the multi-lane overtaking task is subjected to 100 repeated evaluation trials under various traffic densities. Table \ref{table:test_res} presents the statistical results, which reveal that our approach exhibits superior performance in every tested configuration. Notably, even as traffic density increases, the success rate of our method degrades only slightly, which demonstrates that our framework extends the performance boundary of the RL policy.

The ablation study on the orchestrator agent further highlights its core value. We find that without the coordination of this module, the task performance of the policy degraded comprehensively, a phenomenon that is particularly pronounced in high-density traffic. We hypothesize that this is due to an accumulated divergence between the objectives of the reward function and the training curriculum as training progresses. Therefore, we conclude that the dynamic coordination for the training objectives of the orchestrator agent is a key factor in effectively improving policy performance.

Furthermore, statistical test results reveal performance differences among various RL policy learning paradigms. First, compared to baselines that rely solely on human expert knowledge, most methods that leverage LLM and VLM assistance for training generally exhibit superior performance. The performance of the AutoReward, after five iterations of reward generation, significantly outperforms that of the vanilla RL approach. This finding demonstrates the capability of LLMs to generate high-quality reward functions. However, as it only automates the reward generation component, its final performance does not surpass that of Curriculum RL, which is trained with a well-designed manual curriculum. A comparison among methods employing CL techniques further reveals that policies trained with automated design of rewards and curricula comprehensively outperform those using only a manually designed curriculum. Finally, when compared to SOTA methods such as LearningFlow and AutoReward, our framework achieves superior task performance by leveraging multimodal information during the multi-agent collaboration process. This advantage is particularly evident in the multi-lane overtaking tasks with the high-density setting.

We also evaluate the performance evolution of our proposed method at different training stages, specifically within the multi-lane overtaking task under various traffic densities. The statistical results are summarized in Table \ref{table:test_res_evo}, which reveals a clear evolution pattern. In the early stages of training, the RL policy first achieves rapid performance gains in simple tasks with lower density. As training progresses, guided by the adaptive reward functions and curricula dynamically generated by our framework, the performance of policies in more complex tasks also steadily improves. This progressive learning progression indicates that our framework successfully provides appropriate reward-curriculum pairs throughout the entire training process, enabling efficient policy learning through an easy-to-hard progression.

\begin{table*}[]
\renewcommand{\arraystretch}{1}
\tiny % \scriptsize
\centering
\caption{Performance evolution of the proposed method in the multi-Lane overtaking task across different training stages.
}
\label{table:test_res_evo}
\resizebox{\linewidth}{!}{
\begin{tabular}{@{}ccccccccccccc@{}}
\toprule
\multirow{2}{*}{Traning Stage} & \multicolumn{3}{c}{Empty}              & \multicolumn{3}{c}{Low Density}       & \multicolumn{3}{c}{Medium Density}    & \multicolumn{3}{c}{High Density}      \\ \cmidrule(l){2-13} 
                         & SR (\%)       & CR (\%)     & TOR (\%)    & SR (\%)      & CR (\%)     & TOR (\%)    & SR (\%)      & CR (\%)     & TOR (\%)    & SR (\%)      & CR (\%)     & TOR (\%)    \\ \midrule
1              &     100       &     0     &    0       &      92     &      8    &      0     &     63     &     25     &     12      &    45       &     31     &      24     \\
2           &     100      &     0      &     0      &     98      &   2        &     0      &     72      &    18      &     10      &     57      &     30     &     13      \\
3      &      100      &     0      &     0      &      99     &    1      &     0      &     83     &    17     &     0      &     71      &   29       &    0       \\
4      &      100     &      0     &     0      &      98     &     2     &      0     &     91      &     9     &     0      &     85      &   15       &     0      \\
5            &     100     &     0      &     0      &      99     &  1         &      0     &     95      &     5     &     0     &     93      &     7     &  0  \\ \bottomrule
\end{tabular}}
\end{table*}

\begin{table*}[!htbp]
\renewcommand{\arraystretch}{1} 
\scriptsize
\centering
\caption{Generalization performance among different urban driving tasks.
}
\label{table:gen_test_res}
\resizebox{\linewidth}{!}{
\begin{tabular}{@{}cccccccccccccc@{}}
\toprule
\multicolumn{2}{c}{\multirow{2}{*}{Task Types}}                                                    & \multicolumn{3}{c}{Empty}                 & \multicolumn{3}{c}{Low Density} & \multicolumn{3}{c}{Medium Density} & \multicolumn{3}{c}{High Density} \\ \cmidrule(l){3-14} 
\multicolumn{2}{c}{}                                                                               & SR (\%) & CR (\%)                 & TOR (\%) & SR (\%)   & CR (\%)   & TOR (\%)   & SR (\%)    & CR (\%)    & TOR (\%)    & SR (\%)    & CR (\%)   & TOR (\%)   \\ \midrule
\multicolumn{2}{c}{Merging}                                                                        & 100    & 0                      & 0       & 99       & 1        & 0         & 96        & 4         & 0          & 94        & 6        & 0         \\  \midrule
\multirow{3}{*}{\begin{tabular}[c]{@{}c@{}}Unsignalized\\ Intersection\end{tabular}} & Left Turn   & 100    & 0                      & 0       & 97       & 3        & 0         & 93        & 7         & 0          & 90        & 10       & 0         \\
                                                                                     & Go Straight & 100    & 0 & 0       & 100      & 0        & 0         & 95        & 5         & 0          & 91        & 9        & 0         \\
                                                                                     & Right Turn  & 100    & 0                      & 0       & 100      & 0        & 0         & 100       & 0         & 0          & 99        & 1        & 0         \\ \bottomrule
\end{tabular}
}
\end{table*}

\subsection{Generalization Performance in Different Driving Scenarios}

To further validate the generalization capability of our proposed framework, we applied it to two additional and distinct urban driving scenarios for both training and testing, including on-ramp merging and unsignalized intersections. It is worth emphasizing that transferring the framework to these new tasks is highly efficient, as we only need to modify a small amount of scene-specific contextual prompts. All other core training and testing hyperparameters are kept identical to those in the prior multi-lane overtaking scenario.

The statistical results, presented in Table \ref{table:gen_test_res}, demonstrate that our method achieves exceptionally high success rates across these disparate task types and under different traffic densities. This outcome indicates that the proposed framework can effectively handle diverse motion patterns and interaction paradigms of SVs, thereby confirming its superior generalization ability across different types of driving tasks. Crucially, this level of performance is achieved with minimal human effort and time costs for adaptation, highlighting the practicality and scalability of our approach.

\begin{table*}[htbp]
\renewcommand{\arraystretch}{1}
\tiny % \scriptsize
\centering
\caption{Compatibility performance with different RL algorithms in the multi-lane overtaking scenario.
}
\label{table:test_res_rl}
\resizebox{\linewidth}{!}{
\begin{tabular}{@{}clcccccccccccc@{}}
\toprule
\multicolumn{2}{c}{\multirow{2}{*}{Methods}} & \multicolumn{3}{c}{Empty} & \multicolumn{3}{c}{Low Density} & \multicolumn{3}{c}{Medium Density} & \multicolumn{3}{c}{High Density} \\ \cmidrule(l){3-14} 
\multicolumn{2}{c}{}                         & SR (\%) & CR (\%) & TOR (\%) & SR (\%)   & CR (\%)   & TOR (\%)   & SR (\%)    & CR (\%)    & TOR (\%)    & SR (\%)    & CR (\%)   & TOR (\%)   \\ \midrule
\multicolumn{2}{c}{Vanilla DQN}              & 100    & 0      & 0       & 78       & 22       & 0         & 70        & 30        & 0          & 59        & 41       & 0         \\
\multicolumn{2}{c}{OGR DQN}                  & 100    & 0      & 0       & 95       & 5        & 0         & 91        & 9         & 0          & 86        & 14       & 0         \\
\multicolumn{2}{c}{Vanilla SAC}              & 100    & 0      & 0       & 82       & 18       & 0         & 72        & 28        & 0          & 63        & 37       & 0         \\
\multicolumn{2}{c}{OGR SAC}                  & 100    & 0      & 0       & 95       & 5        & 0         & 89        & 11        & 0          & 85        & 15       & 0         \\ 
\multicolumn{2}{c}{Vanilla PPO}              & 99    & 1      & 0       & 80       & 20       & 0         & 69        & 31        & 0          & 62        & 38       & 0         \\
\multicolumn{2}{c}{OGR PPO}                  & 100    & 0      & 0       & 99       & 1        & 0         & 95        & 5         & 0          & 93        & 7       & 0         \\ \bottomrule
\end{tabular}
}
\end{table*}

\subsection{Demonstration of Maneuvering Abilities in Driving Tasks}

To visualize the driving skills acquired through our framework, we present representative test cases for each driving task, illustrating the ability of policies to handle complex interactions. Figs. \ref{fig:sim_demo_overtake}-\ref{fig:sim_demo_intersection} depict the key snapshots of these demonstrations. A detailed description of each demonstration is provided below. 

\subsubsection{Multi-Lane Overtaking Task} 
In Fig. \ref{fig:sim_demo_overtake}, the initial position of the EV is set to the middle lane, with two SVs ahead in the same lane and another SV in the right lane. At 0.0 s, after perceiving the traffic in the middle and right lanes, the RL policy infers that the left lane is suitable for performing an overtaking maneuver. It consequently selects a waypoint to the front-left as a high-level reference and begins to accelerate, initiating a left lane change maneuver. However, the traffic situation evolves dynamically at 1.4 s as the SV in the middle lane closer to the EV suddenly changes lanes to the left. Our policy demonstrates excellent reactive capabilities. It recognizes that sufficient space will open up in the middle lane, and then immediately aborts its initial intent to overtake on the left. Instead, it re-evaluates and decides to back into the middle lane. By 3.2 s, the EV successfully catches up with the two SVs ahead by maintaining a high speed in the middle lane. At 5.9 s, the policy once again assesses the traffic situation. After determining that the left SV is slower than the right one, it again selects a waypoint to the front-left, initiating an attempt at a left lane change to overtake. Finally, at 7.2 s, the EV safely and efficiently completes the multi-lane overtaking task under the guidance of the RL policy.

\subsubsection{On-Ramp Merging Task} 
In Fig. \ref{fig:sim_demo_merging}, the initial position of the EV is set to a random location on the merging lane. A key constraint is introduced by placing a stationary obstacle vehicle at the end of the ramp, requiring the EV to merge into the main carriageway before colliding with the obstacle vehicle. Concurrently, multiple SVs are populated on the main carriageway, traveling at various speeds. During the initial phase, from 0.0 s to 2.8 s, the RL policy selects a nearby waypoint ahead as a reference, guiding the EV to advance along the ramp at a moderate speed while actively seeking a safe gap in the traffic to merge into the main flow. The critical decision-making moment occurs at 5.5 s. At this point, the policy accurately detects that an SV approaching from its rear-right is decelerating and correctly infers this behavior as a yielding intention. Based on this crucial judgment, the policy decisively selects a waypoint to the front-right to initiate the merging maneuver. From 6.1 s to 7.0 s, the EV smoothly completes the maneuver, ultimately integrating safely and efficiently into the traffic on the main carriageway under the continuous guidance of the RL policy.

\subsubsection{Left-Turn Task} 
In Fig. \ref{exp:demo_left}, the EV is initialized in the left lane. At 0.0 s, due to three SVs approaching the center area from three different directions simultaneously, the RL policy guides the EV to drive at a low speed. At 3.0 s, as the SV from the left area maintains its speed while passing through the center area, the RL policy selects a closer waypoint to guide the EV to decelerate and yield. Between 4.2 s and 4.8 s, the SV from the left has already cleared the potential collision area, and the oncoming SV decelerates, exhibiting a yielding intention. Consequently, the RL policy chooses a farther waypoint and a high reference speed to guide the EV rapidly through the intersection. Finally, at 8.7 s, the EV safely and efficiently completes the left-turn task under the guidance of the RL policy.

\subsubsection{Go-Straight Task} 
In Fig. \ref{exp:demo_straight}, the EV is initialized in the right lane. Starting from 0.0 s, the RL policy guides the EV to approach the intersection at a low speed. At 3.0 s, as the SV from the left area maintains a high speed to proceed straight, the RL policy chooses a closer waypoint and a low reference speed to guide the EV to decelerate and yield. Between 4.4 s and 5.8 s, since the SV from the left has already cleared the intersection, the RL policy chooses a farther waypoint and a high reference speed to guide the EV to traverse the intersection. Finally, at 7.3 s, the EV safely and efficiently completes the go-straight task.

\subsubsection{Right-Turn Task} 
In Fig. \ref{exp:demo_right}, the EV is positioned in the left lane. From 0.0 s to 1.7 s, the RL policy first executes a preparatory right lane change and guides the EV to approach the intersection at a low speed, positioning it for the upcoming right turn. At 2.8 s, after assessing the traffic situation, the policy infers that the trajectory of the SV approaching from the left will not conflict with the intended path of the EV. Based on this safety assessment, it begins to guide the EV into the intersection to execute a small-radius right turn. Subsequently, between 3.5 s and 5.1 s, the EV smoothly finishes the right-turn task.

\begin{table}[htbp]
\centering
\caption{Average, maximum, and minimum computation time (s) of the proposed method in various urban scenarios.
}
\begin{tabular}{cccc}
\toprule
           & Average & Maximum & Minimum \\ \midrule
Overtaking &  0.025  &  0.116  &  0.009  \\
Merging    &  0.026  &  0.114  &  0.009  \\
Left-Turn    &  0.027  &  0.116  &  0.010  \\
Go-Straight    &  0.028  &  0.118  &  0.009  \\
Right-Turn    &  0.028  &  0.118  &  0.010  \\ \bottomrule
\end{tabular}
\label{table:compute_time}
\end{table}

\begin{figure*}[htbp]
\centering
    \includegraphics[width=0.95\textwidth]{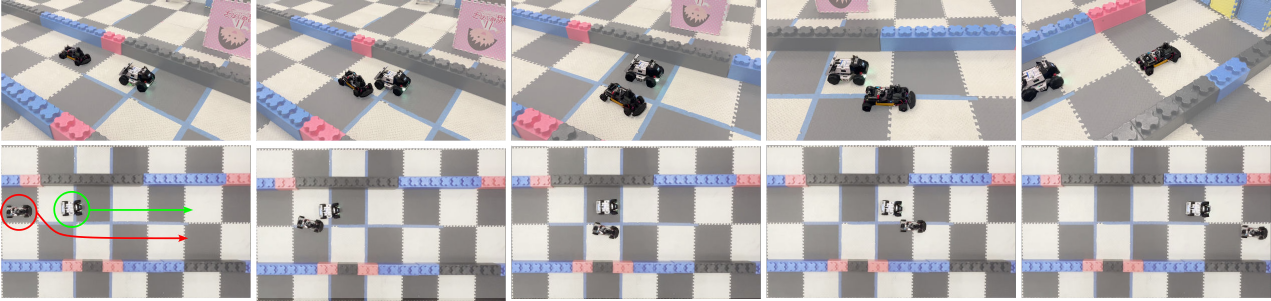}
\caption{Key snapshots of a hardware experiment demonstration in the multi-lane overtaking scenario. In this case, the EV (circled in red) overtakes the low-speed SV (circled in green) ahead by changing lanes to the right. The top sub-figures show the third-person views, while the bottom sub-figures show the corresponding top views. 
The red and green arrows represent the trajectories of the EV and SV, respectively.
}
\label{fig:real_demo_overtake}
\end{figure*}

\begin{figure*}[htbp]
\centering
    \includegraphics[width=0.95\textwidth]{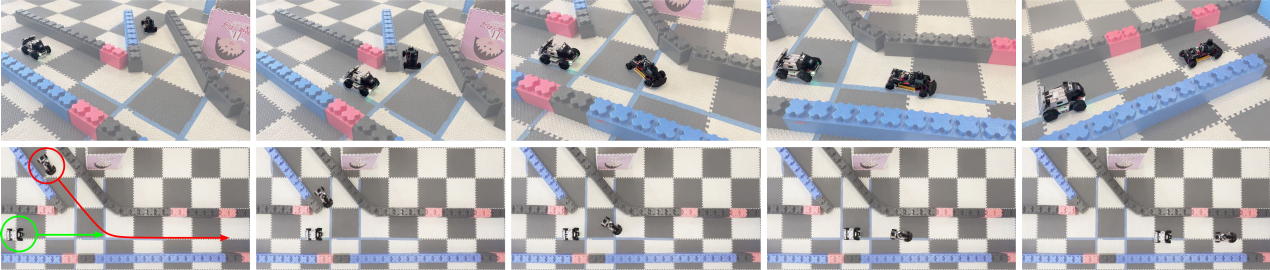}
\caption{Key snapshots of a hardware experiment demonstration in the on-ramp merging scenario. In this case, the EV (circled in red) detects the SV (circled in green) decelerating to yield, and therefore proceeds first to complete the merging task. 
}
\label{fig:real_demo_merging}
\end{figure*}

\begin{figure*}[!htbp]
\centering
    \subfigure[Left-turn task. In this case, the EV detects the SV decelerating to yield and directly performs a left turn to pass through the center area of the intersection.]{\includegraphics[width=0.95\textwidth]{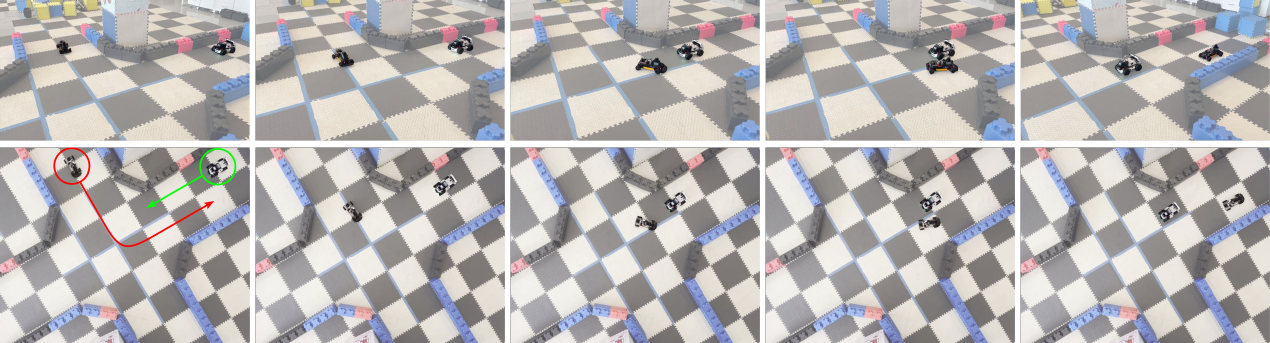}\label{exp:demo_left_real}}
    \subfigure[Go-straight task. In this case, the EV detects that the SV approaching from the left area maintains its speed and goes straight. Therefore, the EV slows down to yield and passes through the center area of the intersection after the SV to complete the go-straight task.]{\includegraphics[width=0.95\textwidth]{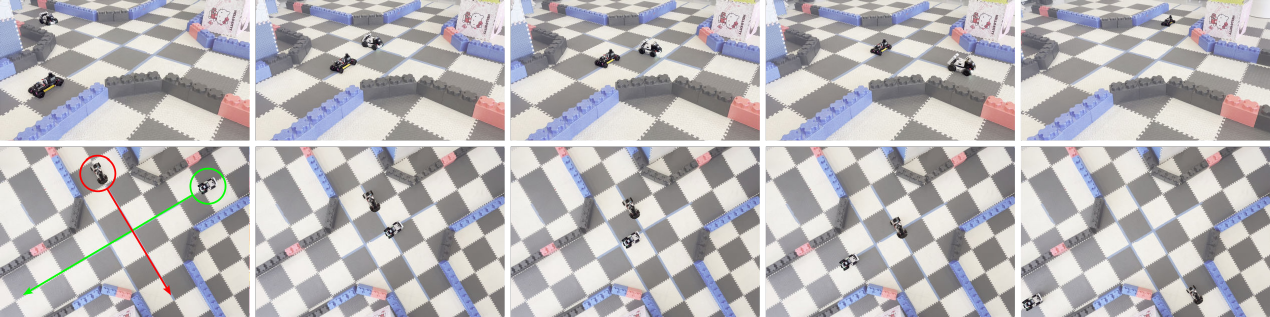}\label{exp:demo_straight_real}}
    \subfigure[Right-turn task. In this case, the EV detects the SV from the left area decelerating to yield and directly performs a right turn to pass through the center area of the intersection.]{\includegraphics[width=0.95\textwidth]{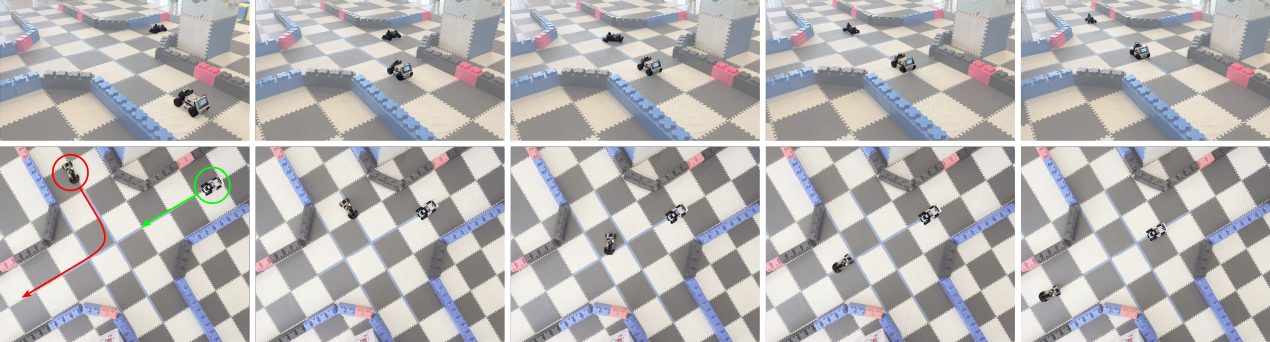}\label{exp:demo_right_real}}
\caption{Key snapshots of three hardware experiment demonstrations in the unsignalized intersection scenario. The EV is highlighted by a red circle and the SV by a green one.
}
\label{fig:real_demo_intersection}
\end{figure*}

To evaluate the real-time performance of our framework, we conduct a statistical analysis of the inference and computation times for the RL executor across different driving tasks. The results are summarized in Table \ref{table:compute_time}. It indicates that the RL executor exhibits stable real-time performance in all driving tasks. Specifically, the maximum computation time for any task is approximately 0.1 seconds, even in the worst case. Furthermore, the results of average computation time demonstrate that the executor consistently operates at a frequency above 30 Hz across all task types. Therefore, the computational efficiency of the RL executor can fully satisfy the real-time requirements for autonomous driving tasks.

\subsection{Compatibility with Different RL Algorithms}

In all preceding experiments, the downstream RL executor is trained using the PPO algorithm. To comprehensively evaluate the compatibility of our framework with different RL algorithms, we select two additional representative algorithms from the RL family, including DQN and SAC. These are used to train policies for the multi-lane overtaking task. To ensure a fair comparison, the relevant hyperparameters for all algorithms are carefully tuned to achieve their respective optimal performance. We then directly compare the policies trained using our framework in conjunction with each RL algorithm against their corresponding vanilla baselines.

The statistical results of all three RL algorithms are summarized in Table \ref{table:test_res_rl}. The quantitative result shows that, regardless of the underlying RL algorithm, policies trained with our integrated framework significantly outperform their vanilla RL counterparts across all test settings. This result indicates that our framework can universally enhance the training process for different types of RL algorithms, thereby substantially improving both sample efficiency and final task performance. Although the performance with DQN and SAC under the proposed framework is slightly inferior to that with PPO, we attribute this discrepancy primarily to the intrinsic adaptability of each algorithm to the high-dimensional continuous and multi-discrete action spaces of this task. Specifically, PPO is generally considered to be more adept in such action spaces. 
Therefore, this result demonstrates that our method exhibits excellent compatibility with a range of mainstream RL algorithms.

\subsection{Real-World Experiments}

To further validate the practical effectiveness of our approach, we carry out a series of real-world experiments. In these experiments, the Tianbot T100, which is equipped with an NVIDIA Jetson Xavier NX, serves as the EV. Meanwhile, the Agilex LIMO platform is employed as the human-operated SV. The experimental setup is implemented using the Robot Operating System (ROS)~\cite{quigley2009ros}, running on a laptop with an 11th Gen Intel(R) Core(TM) i7-1165G7 CPU operating at 2.80 GHz. Vehicle pose information is captured in real time by the OptiTrack motion tracking system at 200 frames per second. For low-level control, a pure pursuit algorithm is applied to follow the high-level decisions generated by RL policies.

Similar to the simulation experiments, we consider three driving scenarios and five driving tasks, including the multi-lane overtaking scenario, on-ramp merging scenario, and unsignalized intersection scenario. Snapshots of these experiments are shown in Figs. \ref{fig:real_demo_overtake}-\ref{fig:real_demo_intersection}. By deploying our trained RL policies in various interactive driving tasks, we find that the EV can make appropriate decisions in response to the diverse intentions exhibited by the SV, thereby completing all driving tasks successfully. Several representative cases illustrate this capability. In the multi-lane overtaking scenario shown in Fig. \ref{fig:real_demo_overtake}, when faced with a lead vehicle maintaining a low speed, the RL policy decisively executes a right lane change and accelerates, completing the overtaking maneuver safely and efficiently. In Fig. \ref{fig:real_demo_merging}, at a critical merging interaction point, the policy infers a yielding intention from a deceleration of the SV. It immediately selects a further waypoint and a high reference speed to seize the opportunity, quickly completing the merge and thus improving traffic efficiency. The three subfigures in Fig. \ref{fig:real_demo_intersection} provide a detailed illustration of how our RL policy flexibly adapts its driving strategy to safely and efficiently complete different tasks based on the various intentions of SVs in the unsignalized intersection scenario. In the left-turn task shown in Fig. \ref{exp:demo_left_real}, the RL policy infers a yielding intention from the SV that is decelerating as it approaches the intersection. Therefore, the policy decisively proceeds before the SV, safely passing through the intersection to efficiently complete the left-turn maneuver. This demonstration shows the ability to seize opportunities of the policy. In contrast, the go-straight task in Fig. \ref{exp:demo_straight_real} showcases the necessary defensive driving behavior of the policy. Recognizing that the SV is maintaining the current velocity, the policy proactively decelerates to yield before entering the conflict zone. After confirming the SV has safely passed, the policy guides the EV to accelerate and cross the intersection. Finally, in the right-turn task depicted in Fig. \ref{exp:demo_right_real}, the policy again correctly interprets the yielding intention of the SV and maintains its heading and speed to smoothly complete the right-turn maneuver.

In conclusion, the above experiments validate the effectiveness of our framework on hardware platforms. The results demonstrate a successful deployment and highlight the superior performance of the proposed method in safe and flexible interaction with SVs of diverse driving intentions across various tasks.

\section{Conclusion}

In this work, we proposed OGR, an advanced automated policy learning framework based on VLM-based multi-agent collaboration. The framework innovatively integrates multiple VLM agents with an RL agent, thereby achieving efficient RL policy training. Specifically, our framework is composed of an orchestration module for training planning, a two-step analyze-then-generate generation module, and an in-training reflection module for online evaluation and selection. To further enhance the robustness of our framework, we integrated parallel generation techniques and a human-in-the-loop reward observation space augmentation mechanism. This multi-agent collaborative system enables the fully automated iterative optimization of reward functions, training curricula, and RL policies. The deep leveraging of multimodal information significantly improves the quality of the generated reward-curriculum pairs, which ultimately translates into superior sample efficiency and task performance of the RL policy. Comprehensive simulation results showcase the effective collaboration among the various agents and demonstrate the superior performance and scalability of our framework on various urban driving tasks and different RL algorithms.Furthermore, real-world experiments validate the potential of our method to be effectively deployed on physical platforms. A promising direction for future work involves leveraging advanced diffusion models to enhance the representational capacity of RL policies.

\bibliographystyle{IEEEtran}
\bibliography{ref}

\begin{thebibliography}{10}
\providecommand{\url}[1]{#1}
\csname url@samestyle\endcsname
\providecommand{\newblock}{\relax}
\providecommand{\bibinfo}[2]{#2}
\providecommand{\BIBentrySTDinterwordspacing}{\spaceskip=0pt\relax}
\providecommand{\BIBentryALTinterwordstretchfactor}{4}
\providecommand{\BIBentryALTinterwordspacing}{\spaceskip=\fontdimen2\font plus
\BIBentryALTinterwordstretchfactor\fontdimen3\font minus \fontdimen4\font\relax}
\providecommand{\BIBforeignlanguage}[2]{{%
\expandafter\ifx\csname l@#1\endcsname\relax
\typeout{** WARNING: IEEEtran.bst: No hyphenation pattern has been}%
\typeout{** loaded for the language `#1'. Using the pattern for}%
\typeout{** the default language instead.}%
\else
\language=\csname l@#1\endcsname
\fi
#2}}
\providecommand{\BIBdecl}{\relax}
\BIBdecl

\bibitem{xu2023multimodal}
P.~Xu, X.~Zhu, and D.~A. Clifton, ``Multimodal learning with transformers: A survey,'' \emph{IEEE Transactions on Pattern Analysis and Machine Intelligence}, vol.~45, no.~10, pp. 12\,113--12\,132, 2023.

\bibitem{hanover2024autonomous}
D.~Hanover, A.~Loquercio, L.~Bauersfeld, A.~Romero, R.~Penicka, Y.~Song, G.~Cioffi, E.~Kaufmann, and D.~Scaramuzza, ``Autonomous drone racing: A survey,'' \emph{IEEE Transactions on Robotics}, vol.~40, pp. 3044--3067, 2024.

\bibitem{han2025multimodal}
X.~Han, S.~Chen, Z.~Fu, Z.~Feng, L.~Fan, D.~An, C.~Wang, L.~Guo, W.~Meng, X.~Zhang \emph{et~al.}, ``Multimodal fusion and vision-language models: A survey for robot vision,'' \emph{arXiv preprint arXiv:2504.02477}, 2025.

\bibitem{zheng2025safe}
L.~Zheng, R.~Yang, M.~Zheng, M.~Y. Wang, and J.~Ma, ``Safe and real-time consistent planning for autonomous vehicles in partially observed environments via parallel consensus optimization,'' \emph{IEEE Transactions on Intelligent Transportation Systems}, pp. 1--17, 2025.

\bibitem{huang2025fast}
Z.~Huang, Y.~Xie, B.~Ma, S.~Shen, and J.~Ma, ``Fast and scalable game-theoretic trajectory planning with intentional uncertainties,'' \emph{arXiv preprint arXiv:2507.12174}, 2025.

\bibitem{cai2023closing}
P.~Cai and D.~Hsu, ``Closing the planning–learning loop with application to autonomous driving,'' \emph{IEEE Transactions on Robotics}, vol.~39, no.~2, pp. 998--1011, 2023.

\bibitem{li2024interactive}
J.~Li, D.~Isele, K.~Lee, J.~Park, K.~Fujimura, and M.~J. Kochenderfer, ``Interactive autonomous navigation with internal state inference and interactivity estimation,'' \emph{IEEE Transactions on Robotics}, vol.~40, pp. 2932--2949, 2024.

\bibitem{zhu2021survey}
Z.~Zhu and H.~Zhao, ``A survey of deep {RL} and {IL} for autonomous driving policy learning,'' \emph{IEEE Transactions on Intelligent Transportation Systems}, vol.~23, no.~9, pp. 14\,043--14\,065, 2021.

\bibitem{chib2023recent}
P.~S. Chib and P.~Singh, ``Recent advancements in end-to-end autonomous driving using deep learning: A survey,'' \emph{IEEE Transactions on Intelligent Vehicles}, vol.~9, no.~1, pp. 103--118, 2023.

\bibitem{cheng2024pluto}
J.~Cheng, Y.~Chen, and Q.~Chen, ``Pluto: Pushing the limit of imitation learning-based planning for autonomous driving,'' \emph{arXiv preprint arXiv:2404.14327}, 2024.

\bibitem{liao2025satp}
H.~Liao, Z.~Li, K.~Zhu, K.~Li, and C.~Xu, ``{SA-TP}$^{2}$: A safety-aware trajectory prediction and planning model for autonomous driving,'' \emph{IEEE Transactions on Robotics}, pp. 1--20, 2025.

\bibitem{ross2011reduction}
S.~Ross, G.~Gordon, and D.~Bagnell, ``A reduction of imitation learning and structured prediction to no-regret online learning,'' in \emph{Proceedings of International Conference on Artificial Intelligence and Statistics}, 2011, pp. 627--635.

\bibitem{zare2024survey}
M.~Zare, P.~M. Kebria, A.~Khosravi, and S.~Nahavandi, ``A survey of imitation learning: Algorithms, recent developments, and challenges,'' \emph{IEEE Transactions on Cybernetics}, vol.~54, no.~12, pp. 7173--7186, 2024.

\bibitem{peng2025bilevel}
Z.~Peng, Y.~Wang, L.~Zheng, and J.~Ma, ``Bilevel multi-armed bandit-based hierarchical reinforcement learning for interaction-aware self-driving at unsignalized intersections,'' \emph{IEEE Transactions on Vehicular Technology}, vol.~74, no.~6, pp. 8824--8838, 2025.

\bibitem{wang2024deep}
X.~Wang, S.~Wang, X.~Liang, D.~Zhao, J.~Huang, X.~Xu, B.~Dai, and Q.~Miao, ``Deep reinforcement learning: A survey,'' \emph{IEEE Transactions on Neural Networks and Learning Systems}, vol.~35, no.~4, pp. 5064--5078, 2024.

\bibitem{biyik2022learning}
E.~B{\i}y{\i}k, D.~P. Losey, M.~Palan, N.~C. Landolfi, G.~Shevchuk, and D.~Sadigh, ``Learning reward functions from diverse sources of human feedback: Optimally integrating demonstrations and preferences,'' \emph{The International Journal of Robotics Research}, vol.~41, no.~1, pp. 45--67, 2022.

\bibitem{yu2025reward}
R.~Yu, S.~Wan, Y.~Wang, C.-X. Gao, L.~Gan, Z.~Zhang, and D.-C. Zhan, ``Reward models in deep reinforcement learning: A survey,'' \emph{arXiv preprint arXiv:2506.15421}, 2025.

\bibitem{yu2018towards}
Y.~Yu, ``Towards sample efficient reinforcement learning,'' in \emph{Proceedings of the International Joint Conference on Artificial Intelligence}.\hskip 1em plus 0.5em minus 0.4em\relax IJCAI, 2018, pp. 5739--5743.

\bibitem{huang2023efficient}
Z.~Huang, J.~Wu, and C.~Lv, ``Efficient deep reinforcement learning with imitative expert priors for autonomous driving,'' \emph{IEEE Transactions on Neural Networks and Learning Systems}, vol.~34, no.~10, pp. 7391--7403, 2023.

\bibitem{hu2020learning}
Y.~Hu, W.~Wang, H.~Jia, Y.~Wang, Y.~Chen, J.~Hao, F.~Wu, and C.~Fan, ``Learning to utilize shaping rewards: A new approach of reward shaping,'' \emph{Advances in Neural Information Processing Systems}, vol.~33, pp. 15\,931--15\,941, 2020.

\bibitem{devidze2022exploration}
R.~Devidze, P.~Kamalaruban, and A.~Singla, ``Exploration-guided reward shaping for reinforcement learning under sparse rewards,'' \emph{Advances in Neural Information Processing Systems}, vol.~35, pp. 5829--5842, 2022.

\bibitem{narvekar2020curriculum}
S.~Narvekar, B.~Peng, M.~Leonetti, J.~Sinapov, M.~E. Taylor, and P.~Stone, ``Curriculum learning for reinforcement learning domains: A framework and survey,'' \emph{Journal of Machine Learning Research}, vol.~21, no. 181, pp. 1--50, 2020.

\bibitem{wang2022surveycl}
X.~Wang, Y.~Chen, and W.~Zhu, ``A survey on curriculum learning,'' \emph{IEEE Transactions on Pattern Analysis and Machine Intelligence}, vol.~44, no.~9, pp. 4555--4576, 2022.

\bibitem{zhang2024vision}
J.~Zhang, J.~Huang, S.~Jin, and S.~Lu, ``Vision-language models for vision tasks: A survey,'' \emph{IEEE Transactions on Pattern Analysis and Machine Intelligence}, vol.~46, no.~8, pp. 5625--5644, 2024.

\bibitem{yin2024survey}
S.~Yin, C.~Fu, S.~Zhao, K.~Li, X.~Sun, T.~Xu, and E.~Chen, ``A survey on multimodal large language models,'' \emph{National Science Review}, vol.~11, no.~12, pp. 1--20, 2024.

\bibitem{tiandrivevlm}
X.~Tian, J.~Gu, B.~Li, Y.~Liu, Y.~Wang, Z.~Zhao, K.~Zhan, P.~Jia, X.~Lang, and H.~Zhao, ``Drivevlm: The convergence of autonomous driving and large vision-language models,'' in \emph{Proceedings of the Conference on Robot Learning}, 2024.

\bibitem{sima2024drivelm}
C.~Sima, K.~Renz, K.~Chitta, L.~Chen, H.~Zhang, C.~Xie, J.~Bei{\ss}wenger, P.~Luo, A.~Geiger, and H.~Li, ``{DriveLM: Driving with graph visual question answering},'' in \emph{European Conference on Computer Vision}.\hskip 1em plus 0.5em minus 0.4em\relax Springer, 2024, pp. 256--274.

\bibitem{guo2024redcode}
C.~Guo, X.~Liu, C.~Xie, A.~Zhou, Y.~Zeng, Z.~Lin, D.~Song, and B.~Li, ``Redcode: Risky code execution and generation benchmark for code agents,'' \emph{Advances in Neural Information Processing Systems}, vol.~37, pp. 106\,190--106\,236, 2024.

\bibitem{wang2024voyager}
G.~Wang, Y.~Xie, Y.~Jiang, A.~Mandlekar, C.~Xiao, Y.~Zhu, L.~Fan, and A.~Anandkumar, ``Voyager: An open-ended embodied agent with large language models,'' \emph{Transactions on Machine Learning Research}, pp. 1--21, 2024.

\bibitem{li2023camel}
G.~Li, H.~Hammoud, H.~Itani, D.~Khizbullin, and B.~Ghanem, ``Camel: Communicative agents for {``mind''} exploration of large language model society,'' \emph{Advances in Neural Information Processing Systems}, vol.~36, pp. 51\,991--52\,008, 2023.

\bibitem{hu2025owl}
M.~Hu, Y.~Zhou, W.~Fan, Y.~Nie, B.~Xia, T.~Sun, Z.~Ye, Z.~Jin, Y.~Li, Q.~Chen \emph{et~al.}, ``Owl: Optimized workforce learning for general multi-agent assistance in real-world task automation,'' \emph{arXiv preprint arXiv:2505.23885}, 2025.

\bibitem{hong2024metagpt}
S.~Hong, M.~Zhuge, J.~Chen, X.~Zheng, Y.~Cheng, C.~Zhang, J.~Wang, Z.~Wang, S.~K.~S. Yau, Z.~Lin \emph{et~al.}, ``{MetaGPT: Meta Programming for A Multi-Agent Collaborative Framework},'' in \emph{International Conference on Learning Representations}, 2024.

\bibitem{zhao2024surveyad}
R.~Zhao, Y.~Li, Y.~Fan, F.~Gao, M.~Tsukada, and Z.~Gao, ``A survey on recent advancements in autonomous driving using deep reinforcement learning: Applications, challenges, and solutions,'' \emph{IEEE Transactions on Intelligent Transportation Systems}, vol.~25, no.~12, pp. 19\,365--19\,398, 2024.

\bibitem{al2023self}
M.~Al-Sharman, R.~Dempster, M.~A. Daoud, M.~Nasr, D.~Rayside, and W.~Melek, ``Self-learned autonomous driving at unsignalized intersections: A hierarchical reinforced learning approach for feasible decision-making,'' \emph{IEEE Transactions on Intelligent Transportation Systems}, vol.~24, no.~11, pp. 12\,345--12\,356, 2023.

\bibitem{wang2023learning}
Y.~Wang, Z.~Peng, Y.~Xie, Y.~Li, H.~Ghazzai, and J.~Ma, ``Learning the references of online model predictive control for urban self-driving,'' \emph{arXiv preprint arXiv:2308.15808}, 2023.

\bibitem{chen2022interpretable}
J.~Chen, S.~E. Li, and M.~Tomizuka, ``Interpretable end-to-end urban autonomous driving with latent deep reinforcement learning,'' \emph{IEEE Transactions on Intelligent Transportation Systems}, vol.~23, no.~6, pp. 5068--5078, 2022.

\bibitem{chen2024risk}
D.~Chen, H.~Li, Z.~Jin, H.~Tu, and M.~Zhu, ``Risk-anticipatory autonomous driving strategies considering vehicles’ weights based on hierarchical deep reinforcement learning,'' \emph{IEEE Transactions on Intelligent Transportation Systems}, vol.~25, no.~12, pp. 19\,605--19\,618, 2024.

\bibitem{abouelazm2024review}
A.~Abouelazm, J.~Michel, and J.~M. Z{\"o}llner, ``A review of reward functions for reinforcement learning in the context of autonomous driving,'' in \emph{IEEE Intelligent Vehicles Symposium}, 2024, pp. 156--163.

\bibitem{golchoubian2024uncertainty}
M.~Golchoubian, M.~Ghafurian, K.~Dautenhahn, and N.~L. Azad, ``Uncertainty-aware drl for autonomous vehicle crowd navigation in shared space,'' \emph{IEEE Transactions on Intelligent Vehicles}, vol.~9, no.~12, pp. 7931--7944, 2024.

\bibitem{huang2023conditional}
Z.~Huang, H.~Liu, J.~Wu, and C.~Lv, ``Conditional predictive behavior planning with inverse reinforcement learning for human-like autonomous driving,'' \emph{IEEE Transactions on Intelligent Transportation Systems}, vol.~24, no.~7, pp. 7244--7258, 2023.

\bibitem{li2024simulation}
W.~Li, F.~Qiu, L.~Li, Y.~Zhang, and K.~Wang, ``Simulation of vehicle interaction behavior in merging scenarios: A deep maximum entropy-inverse reinforcement learning method combined with game theory,'' \emph{IEEE Transactions on Intelligent Vehicles}, vol.~9, no.~1, pp. 1079--1093, 2024.

\bibitem{guan2020centralized}
Y.~Guan, Y.~Ren, S.~E. Li, Q.~Sun, L.~Luo, and K.~Li, ``Centralized cooperation for connected and automated vehicles at intersections by proximal policy optimization,'' \emph{IEEE Transactions on Vehicular Technology}, vol.~69, no.~11, pp. 12\,597--12\,608, 2020.

\bibitem{song2021autonomous}
Y.~Song, H.~Lin, E.~Kaufmann, P.~D{\"u}rr, and D.~Scaramuzza, ``Autonomous overtaking in {Gran} {Turismo} sport using curriculum reinforcement learning,'' in \emph{IEEE International Conference on Robotics and Automation}, 2021, pp. 9403--9409.

\bibitem{khaitan2022state}
S.~Khaitan and J.~M. Dolan, ``State dropout-based curriculum reinforcement learning for self-driving at unsignalized intersections,'' in \emph{IEEE/RSJ International Conference on Intelligent Robots and Systems}, 2022, pp. 12\,219--12\,224.

\bibitem{peng2023CPPO}
Z.~Peng, X.~Zhou, Y.~Wang, L.~Zheng, M.~Liu, and J.~Ma, ``Curriculum proximal policy optimization with stage-decaying clipping for self-driving at unsignalized intersections,'' in \emph{Proceedings of the International Intelligent Transportation Systems Conference}, 2023, pp. 5027--5033.

\bibitem{peng2024reward}
Z.~Peng, X.~Zhou, L.~Zheng, Y.~Wang, and J.~Ma, ``Reward-driven automated curriculum learning for interaction-aware self-driving at unsignalized intersections,'' in \emph{IEEE/RSJ International Conference on Intelligent Robots and Systems}, 2024.

\bibitem{mei2024continuously}
J.~Mei, Y.~Ma, X.~Yang, L.~Wen, X.~Cai, X.~Li, D.~Fu, B.~Zhang, P.~Cai, M.~Dou \emph{et~al.}, ``Continuously learning, adapting, and improving: A dual-process approach to autonomous driving,'' \emph{Advances in Neural Information Processing Systems}, vol.~37, pp. 123\,261--123\,290, 2024.

\bibitem{xu2024drivegpt4}
Z.~Xu, Y.~Zhang, E.~Xie, Z.~Zhao, Y.~Guo, K.-Y.~K. Wong, Z.~Li, and H.~Zhao, ``{DriveGPT}4: Interpretable end-to-end autonomous driving via large language model,'' \emph{IEEE Robotics and Automation Letters}, vol.~9, no.~10, pp. 8186--8193, 2024.

\bibitem{shao2024lmdrive}
H.~Shao, Y.~Hu, L.~Wang, G.~Song, S.~L. Waslander, Y.~Liu, and H.~Li, ``{LMD}rive: Closed-loop end-to-end driving with large language models,'' in \emph{Proceedings of the IEEE/CVF Conference on Computer Vision and Pattern Recognition}, 2024, pp. 15\,120--15\,130.

\bibitem{wen2023dilu}
L.~Wen, D.~Fu, X.~Li, X.~Cai, T.~Ma, P.~Cai, M.~Dou, B.~Shi, L.~He, and Y.~Qiao, ``{DiLu: A knowledge-driven approach to autonomous driving with large language models},'' in \emph{Proceedings of International Conference on Learning Representations}, 2024, pp. 1--20.

\bibitem{ye2024lord}
X.~Ye, F.~Tao, A.~Mallik, B.~Yaman, and L.~Ren, ``{LORD}: Large models based opposite reward design for autonomous driving,'' in \emph{IEEE/CVF Winter Conference on Applications of Computer Vision}, 2025, pp. 5072--5081.

\bibitem{huang2024vlm}
Z.~Huang, Z.~Sheng, Y.~Qu, J.~You, and S.~Chen, ``{VLM}-{RL}: A unified vision language models and reinforcement learning framework for safe autonomous driving,'' \emph{arXiv preprint arXiv:2412.15544}, 2024.

\bibitem{ma2024eureka}
Y.~J. Ma, W.~Liang, G.~Wang, D.-A. Huang, O.~Bastani, D.~Jayaraman, Y.~Zhu, L.~Fan, and A.~Anandkumar, ``Eureka: Human-level reward design via coding large language models,'' in \emph{Proceedings of International Conference on Learning Representations}, 2024, pp. 1--14.

\bibitem{han2024autoreward}
X.~Han, Q.~Yang, X.~Chen, Z.~Cai, X.~Chu, and M.~Zhu, ``Autoreward: Closed-loop reward design with large language models for autonomous driving,'' \emph{IEEE Transactions on Intelligent Vehicles}, pp. 1--13, 2024.

\bibitem{li2024sds}
J.~Li, M.~Stamatopoulou, and D.~Kanoulas, ``{SDS}--see it, do it, sorted: Quadruped skill synthesis from single video demonstration,'' \emph{arXiv preprint arXiv:2410.11571}, 2024.

\bibitem{liang2024eurekaverse}
W.~Liang, S.~Wang, H.-J. Wang, O.~Bastani, D.~Jayaraman, and Y.~J. Ma, ``Environment curriculum generation via large language models,'' in \emph{Proceedings of the Conference on Robot Learning}, 2024, pp. 1--14.

\bibitem{ryu2024curricullm}
K.~Ryu, Q.~Liao, Z.~Li, K.~Sreenath, and N.~Mehr, ``Curricullm: Automatic task curricula design for learning complex robot skills using large language models,'' \emph{arXiv preprint arXiv:2409.18382}, 2024.

\bibitem{sheng2025curricuvlm}
Z.~Sheng, Z.~Huang, Y.~Qu, Y.~Leng, S.~Bhavanam, and S.~Chen, ``Curricuvlm: Towards safe autonomous driving via personalized safety-critical curriculum learning with vision-language models,'' \emph{arXiv preprint arXiv:2502.15119}, 2025.

\bibitem{peng2025learningflow}
Z.~Peng, Y.~Wang, X.~Han, L.~Zheng, and J.~Ma, ``Learning{F}low: Automated policy learning workflow for urban driving with large language models,'' \emph{arXiv preprint arXiv:2501.05057}, 2025.

\bibitem{singh2009rewards}
S.~Singh, R.~L. Lewis, and A.~G. Barto, ``Where do rewards come from,'' in \emph{Proceedings of the Annual Conference of the Cognitive Science Society}.\hskip 1em plus 0.5em minus 0.4em\relax Cognitive Science Society, 2009, pp. 2601--2606.

\bibitem{huang2023agentcoder}
D.~Huang, J.~M. Zhang, M.~Luck, Q.~Bu, Y.~Qing, and H.~Cui, ``Agentcoder: Multi-agent-based code generation with iterative testing and optimisation,'' \emph{arXiv preprint arXiv:2312.13010}, 2023.

\bibitem{zhang2020automatic}
Y.~Zhang, P.~Abbeel, and L.~Pinto, ``Automatic curriculum learning through value disagreement,'' \emph{Advances in Neural Information Processing Systems}, vol.~33, pp. 7648--7659, 2020.

\bibitem{tao2024reverse}
S.~Tao, A.~Shukla, T.-k. Chan, and H.~Su, ``Reverse forward curriculum learning for extreme sample and demonstration efficiency in reinforcement learning,'' in \emph{Proceedings of International Conference on Learning Representations}, 2024, pp. 1--13.

\bibitem{dosovitskiy2017carla}
A.~Dosovitskiy, G.~Ros, F.~Codevilla, A.~Lopez, and V.~Koltun, ``{CARLA}: An open urban driving simulator,'' in \emph{Proceedings of the Conference on Robot Learning}, 2017, pp. 1--16.

\bibitem{schulman2017proximal}
J.~Schulman, F.~Wolski, P.~Dhariwal, A.~Radford, and O.~Klimov, ``Proximal policy optimization algorithms,'' \emph{arXiv preprint arXiv:1707.06347}, 2017.

\bibitem{kingma2014adam}
D.~P. Kingma and J.~Ba, ``Adam: A method for stochastic optimization,'' \emph{arXiv preprint arXiv:1412.6980}, 2014.

\bibitem{andersson2019casadi}
J.~A. Andersson, J.~Gillis, G.~Horn, J.~B. Rawlings, and M.~Diehl, ``Casadi: a software framework for nonlinear optimization and optimal control,'' \emph{Mathematical Programming Computation}, vol.~11, pp. 1--36, 2019.

\bibitem{quigley2009ros}
M.~Quigley, K.~Conley, B.~Gerkey, J.~Faust, T.~Foote, J.~Leibs, R.~Wheeler, A.~Y. Ng \emph{et~al.}, ``{{ROS}}: an open-source robot operating system,'' in \emph{{ICRA Workshop on Open Source Software}}, vol.~3, no. 3.2.\hskip 1em plus 0.5em minus 0.4em\relax Kobe, 2009, p.~5.

\end{thebibliography}

\vfill

\end{document}